\documentclass{article}
\usepackage{graphicx}

\usepackage{amsmath}
\usepackage{amssymb}
\usepackage{amsfonts}
\usepackage[linesnumbered,algoruled,lined]{algorithm2e}
\usepackage{multirow}
\usepackage{float}
\usepackage{subfigure}
\usepackage{bbm}

\usepackage{hyperref}
\hypersetup{
	colorlinks=true,
	linkcolor=blue,
	citecolor=blue,
	filecolor=blue,      
	urlcolor=blue,
}
\allowdisplaybreaks

\newcommand{\eproof}{\hfill\rule{2.2mm}{3.0mm}}

\newcommand{\Proof}{\noindent {\bf Proof.~~}}

\newcommand{\R}{{\mathbb R}}

\renewcommand{\eqref}[1]{(\ref{#1})}

\numberwithin{equation}{section}

\newcommand{\rank}{{\rm rank}}
\newcommand{\diag}{{\rm diag}}
\newcommand{\tr}{{\rm tr}}

\newcommand{\vn}{{\mathbf n}}
\renewcommand{\R}{{\mathbb R}}

\newtheorem{theo}{Theorem}[section]
\newtheorem{lem}[theo]{Lemma}

\newtheorem{definition}{Definition}[section]
\newcommand\keywords{\noindent\textbf{Keywords:}\ }

\title{The Exploration of Neural Collapse under Imbalanced Data}
\author{Haixia Liu\thanks{School of Mathematics and Statistics  \& Institute of Interdisciplinary Research for Mathematics and Applied Science \& Hubei Key Laboratory of Engineering Modeling and Scientific Computing, Huazhong University of Science and Technology, Wuhan, Hubei, China. Email: liuhaixia@hust.edu.cn. The work of H.X. Liu was supported in part by NSFC 11901220, Interdisciplinary Research Program of HUST 2024JCYJ005, National Key Research and Development Program of China 2023YFC3804500 and Hubei Key Laboratory of Engineering Modeling and Scientific Computing.}} 
\begin{document}
\maketitle


%
%

\begin{abstract}
Neural collapse, a newly identified characteristic, describes a property of solutions during model training. In this paper, we explore neural collapse in the context of imbalanced data. We consider the 
$L$-extended unconstrained feature model with a bias term and provide a theoretical analysis of global minimizer.

Our findings include: (1) Features within the same class converge to their class mean, similar to both the balanced case and the imbalanced case without bias. (2) The geometric structure is mainly on the left orthonormal transformation of the product of $L$ linear classifiers and the right transformation of the class-mean matrix. (3) Some rows of the left orthonormal transformation of the product of $L$ linear classifiers collapse to zeros and others are orthogonal, which relies on the singular values of $\hat Y=(I_K-1/N\vn\mathbbm{1}^\top_K)D$, where $K$ is class size, $\vn$ is the vector of sample size for each class, $D$ is the diagonal matrix whose diagonal entries are given by $\sqrt{\vn}$. Similar results are for the columns of the right orthonormal transformation of the product of class-mean matrix and $D$. (4) The $i$-th row of the left orthonormal transformation of the product of $L$ linear classifiers aligns with the $i$-th column of the right orthonormal transformation of the product of class-mean matrix and $D$. (5) We provide the estimation of singular values about $\hat Y$. Our numerical experiments support these theoretical findings.

\end{abstract}





\section{Introduction}
\label{sec:intro}
In recent years, deep neural networks have been widely applied in various fields such as speech recognition, image analysis, semantic segmentation, face recognition, object detection, and autonomous driving, due to their outstanding algorithmic performance \cite{deng2013new,litjens2017survey,garcia2017review,hu2015face,zhao2019object,grigorescu2020survey}. The reason why deep neural networks have achieved such tremendous success is that they emulate the learning system of the human brain, enabling machines to learn abstract features from data through deep layers of neural networks. Despite the widespread application of modern deep learning methods in various domains, the mysteries behind them are not fully understood. The practice of deep learning involves appropriate network architecture design, as well as the generalization capability \cite{qi2020deep,martens2021rapid} and robustness \cite{nakkiran2021deep,yang2020rethinking,madry2018towards} of the learned networks, which are shrouded in mystery. In order to unravel the mysteries of deep neural networks, some research studies \cite{papyan2020prevalence,fang2021exploring,graf2021dissecting,han2021neural,ji2021unconstrained,lu2022neural,mixon2020neural,tirer2022extended,zhu2021geometric,zhou2022optimization} have focused on the final stage of deep representation learning during the training process and have discovered some useful phenomena. In particular, recent groundbreaking work \cite{papyan2020prevalence} has found in numerical experiments that there exists a characteristic called Neural Collapse (NC) between the last layer features of well-trained deep neural networks (outputs of the penultimate layer) and the classifier, including
\begin{itemize}
    \item {\bf Variability collapse:} the individual features of each class concentrate to their class-means.
    \item {\bf Convergence to simplex ETF:} the class-means have the same length and are maximally distant; they form a Simplex Equiangular Tight Frame (ETF).
    \item {\bf Convergence to self-duality:} the last-layer linear classifiers perfectly match their class-means.
    \item {\bf Simple decision rule:} the last-layer classifier is equivalent to a Nearest Class-Center decision rule.
\end{itemize}

The result of neural collapse indicates that the learning objective of deep neural networks is to maximize the separability between different classes, allowing the classifier to achieve maximum margin classification and reach the limits of training performance. However, the aforementioned simple geometric structure is derived and validated based on datasets with balanced data (equal number of samples per class). In practical applications, it is rare to have completely balanced datasets across classes. Imbalances among different classes is a common occurrence in various learning (training) tasks. For example, {\bf Medical diagnosis:} In the field of medicine, the occurrence rate of certain rare diseases can be very low, resulting in a significantly smaller number of samples for rare diseases compared to common diseases. 
{\bf Fraud detection:} In credit card fraud detection tasks, the number of normal transactions is typically much higher than fraudulent transactions. As a result, the number of samples for fraudulent transactions is relatively small, leading to class imbalance. 
{\bf Text classification:} In natural language processing tasks, such as sentiment analysis, there may be a noticeable difference in the number of positive and negative texts. For example, on social media, positive comments may be more common than negative comments. 
{\bf Defect detection:} In manufacturing industries, some products may have a low defect ratio, resulting in a significantly higher number of samples for normal products compared to defective products. 

In such cases, the classifier may perform better on classes with a larger number of samples, while exhibiting higher misclassification rates on classes with fewer samples, which disrupts the symmetrical geometric structure. Moreover, the theoretical research on neural collapse phenomenon is mostly derived and validated under the assumption of balanced data, with limited theoretical results addressing the scenario under imbalanced data.

Although previous research has made some progress in studying the neural collapse phenomenon under imbalanced class distributions in datasets, these theories are based on certain specific conditions. For instance, \cite{thrampoulidis2022imbalance} only considers the Hinge loss function with a bias term of 0 and assumes that the dataset is divided into two classes based on the sample count: majority class and minority class. Fang et al. \cite{fang2021exploring} analyze the Layer-Peeled model in the context of class imbalances, revealing phenomena such as minority collapse that hinder the performance of deep learning models. More recently, Dang et al. \cite{dang2023neural} have extended the theory of neural collapse to a bias-free $L$-layer framework, addressing imbalanced datasets. Hong and Ling \cite{hong2024neural} study the extension of NC phenomenon to
imbalanced datasets under cross-entropy loss function in the context of the unconstrained
feature model. However, to date, there are still several gaps in the research on the neural collapse phenomenon under imbalanced data, for example, the lack of theoretical research on unconstrained feature models or optimization models with multiple layers with non-zero bias terms.

In this paper, we investigate the global optimal solutions of the Unconstrained Feature Model (UFM) under both bias-free and biased conditions in the context of imbalanced data, and subsequently extend our analysis to an 
$L$-layer scenario. Our findings include: (1) Features within the same class converge to their class mean, similar to both the balanced case and the imbalanced case without bias. (2) The geometric structure is mainly on the left orthonormal transformation of the product of $L$ linear classifiers and the right transformation of the class-mean matrix. (3) Some rows of the left orthonormal transformation of the product of $L$ linear classifiers collapse to zeros and others are orthogonal, which relies on the singular values of $\hat Y=(I_K-1/N\vn\mathbbm{1}^\top_K)D$, where $K$ is class size, $\vn$ is the vector of sample size for each class, $D$ is the diagonal matrix whose diagonal entries are given by $\sqrt{\vn}$. Similar results are for the columns of the right orthonormal transformation of the product of class-mean matrix and $D$. (4) The $i$-th row of the left orthonormal transformation of the product of $L$ linear classifiers aligns with the $i$-th column of the right orthonormal transformation of the product of class-mean matrix and $D$. (5) We provide the estimation of singular values about $\hat Y$. (6) We assess our theoretical findings through extensive numerical evaluations.

The rest of this paper is organized as follows. Section \ref{sec:related} reviews the related works about neural collapse. Section \ref{sec:problem} presents the problem setup, including the definitions of imbalanced data and the unconstrained feature model, which is followed by the theoretical analysis about the unconstrained feature model on both bias-free and bias cases under the imbalanced data and the extended results to $L$ layers in Section \ref{sec:theory}. Numerical results and some technical details of proof are provided in Sections \ref{sec:numerical} and \ref{sec:details}. Finally, we conclude the results in \ref{sec:conclusion}.
\section{Related Works}
\label{sec:related}
In 2020, Papyan, Han, and Donoho were the first researchers to introduce the term ``neural collapse'' to describe a property of solutions during model training. They provided a detailed characterization of solution properties based on four behaviors of classifiers and the last layer features \cite{papyan2020prevalence}. This research has had a significant impact on the solution of deep learning problems, offering valuable insights for better understanding and improving the model training process. Since then, in a short period of three to four years, many scholars have started to focus on and study this issue. Recent research efforts have been dedicated to analyzing the behavior of neural collapse in deep neural networks from a theoretical perspective \cite{fang2021exploring,graf2021dissecting,han2021neural,ji2021unconstrained,lu2022neural,mixon2020neural,tirer2022extended,zhu2021geometric,zhou2022optimization}.

Firstly, Mixon et al. \cite{mixon2020neural} proposed a simplified mathematical framework called the Unconstrained Feature Model (UFM) for theoretical analysis. The basic principle of this framework is that modern deep networks are overparameterized and expressive \cite{gybenko1989approximation,hornik1991approximation,lu2017expressive,shaham2018provable}, and their feature mappings can fit any training data. Therefore, the features in the last layer can be treated as free variables. Subsequent theoretical analyses have mostly been built upon this mathematical framework \cite{fang2021exploring,lu2022neural,tirer2022extended,weinan2022emergence}. Under the Unconstrained Feature Model, related studies \cite{graf2021dissecting,ji2021unconstrained,lu2022neural,mixon2020neural,tirer2022extended,weinan2022emergence} indicate that the Neural Collapse (NC) solution is the unique global optimum. Thus, in the Unconstrained Feature Model, despite the non-convexity of the optimization objective function, it is still possible to guarantee reaching the global optimum. Recently, Zhu et al. \cite{zhu2021geometric} further addressed this issue by showing that the cross-entropy loss function exhibits a favorable global optimization landscape under the Unconstrained Feature Model. The research results indicate that each saddle point is a strict saddle point with negative curvature. Therefore, regardless of the non-convexity, the NC solution can be efficiently optimized through the cross-entropy loss function.

Previous research has mostly focused on the cross-entropy (CE) loss function, but the mean squared error (MSE) loss function has attracted widespread attention due to its simple algebraic expression and excellent performance in training deep neural networks \cite{hui2020evaluation}. Han et al. \cite{han2021neural} studied the neural collapse phenomenon under the MSE loss function. Assuming the classifier is a least squares classifier, the authors first decomposed the MSE loss function into a least squares part and a least squares bias part. Numerical results showed that the bias part can be ignored during the training process, leading to the introduction of the theoretical construct called the central path. The study found that on the central path, linear classifiers maintain optimality of the MSE loss function throughout the dynamic process. Further analysis of the refined gradient flow along the central path accurately predicted the neural collapse phenomenon. Tirer et al. \cite{tirer2022extended} investigated the global optimality conditions for the neural collapse phenomenon exhibited by the MSE loss function in two-layer and three-layer networks, including special cases with no bias term or weight decay. Zhou et al. \cite{zhou2022optimization} conducted a deep landscape analysis of the last layer features in deep neural networks and proved that under the Unconstrained Feature Model, the stationary points are either global optima or strict saddle points.

The aforementioned theoretical analyses are primarily based on balanced datasets, with relatively fewer theoretical analyses addressing the neural collapse 
 on imbalanced data \cite{fang2021exploring,thrampoulidis2022imbalance,dang2023neural}. Fang et al. \cite{fang2021exploring} proposed the concept of Minority Collapse phenomenon in the case of imbalanced data. They theoretically demonstrated the limited performance of deep learning models on minority class categories. Thrampoulidis et al. \cite{thrampoulidis2022imbalance} conducted geometric analysis of neural collapse under imbalanced data. They used the Unconstrained Feature Model and the Hinge loss function to propose a method called Simplex-Encoded-Labels Interpolation (SELI) to characterize the neural collapse phenomenon. The study found that the last layer features and classifiers always interpolate a simplex-encoded label matrix, and the geometric structure of the last layer features and classifiers is determined by the singular value decomposition (SVD) factors of that matrix. More recently, Dang et al. \cite{dang2023neural} have extended the theory of neural collapse to a bias-free $L$-layer framework, addressing imbalanced datasets. Hong and Ling \cite{hong2024neural} study the extension of NC phenomenon to
imbalanced datasets under cross-entropy loss function in the context of the unconstrained
feature model. 
\section{Problem Setup}
\label{sec:problem}
The task of deep learning is to find a map in the training data, which map the inputs to the corresponding labels, and then generalize well on the data out of the training data. Let $x$ be the input, $\phi_\theta(x)$ be the feature map, $(W,b)$ be the parameters of linear classifier, then the goal is to find the map from the inputs to the corresponding labels
\[\psi_\Theta(x):=W\phi_\theta(x)+b,\]
where $\Theta=(\theta,W,b)$.

In this work, we focus on a multi-class task in classification. Assume there are $K$ classes with $n_k$ samples in the $k$-th class for $k=1,\cdots,K$, then the total number of samples are $N=\sum^K_{k=1}n_k$. Let $x_{k,i}$ be the $i$-th sample in the $k$-th class and the corresponding label vector is a one-hot vector, denoted as $y_k$ for $i=1,\cdots,n_k$ and $k=1,\cdots,K$. 

In training phase, our task is to learning the parameter $\Theta$ such that the output of classifier is as close to the corresponding label as possible. Here we consider the regularized MSE model to measure the differences between the output and the corresponding label. The optimization model is
\[\min_{\Theta}\frac{1}{2N}\sum^K_{k=1}\sum^{n_k}_{i=1}\|\psi_\Theta(x_{k,i})-y_k\|^2_2+\rho\|\Theta\|^2_F,\]
where $\rho\ge0$ is the regularized parameter. 

In general, it is extremely difficult to do theoretical analysis of deep neural networks as the layer goes to deeper. Fortunately, modern deep networks are often highly overparameterized to approximate any continuous function \cite{gybenko1989approximation,hornik1991approximation,lu2017expressive,shaham2018provable}. To simplify theoretical analysis as \cite{mixon2020neural,zhou2022optimization,zhu2021geometric}, we focus on the $L$-extended unconstrained feature model, which consider the outputs of the last $L$ layer(s) as free variables in the following. When $L=1$, it degrades to the unconstrained feature model.
\begin{definition}[$L$-extended unconstrained feature model]
    \label{def:ufm}
    Let $W_i\in\R^{d_i\times d_{i-1}},i=1,\cdots,L-1$, $W_L\in\R^{K\times d_{L-1}}$, $H\in\R^{d_0\times N}$, $b\in\R^{K}$ and $Y\in\R^{K\times N}$, where each column of $Y$ is a one-hot vector. Then the $L$-Extended Unconstrained Feature Model ($L$-EUFM) is
    \begin{equation*}
\begin{split}
    &\min_{W_L,\cdots,W_1,H,b}\ f(W_L,\cdots,W_1,H,b)\\
    =&\min_{W_L,\cdots,W_1,H,b}\ \frac1{2N}\|W_L\cdots W_1H+b\mathbbm{1}^{\top}_N-Y\|^2_F+\frac{1}{2}\sum^L_{j=1}\lambda_{W_j}\|W_j\|^2_F+\frac{\lambda_{H}}2\|H\|^2_F+\frac{\rho_b}{2}\|b\|^2_2,
\end{split}
\end{equation*}
    where $\rho_{W_i}\ge0$, $\rho_H\ge0$ and $\rho_b\ge 0$ for $i=1,\cdots,L$. Throughout of this paper, we set $\rho_b=0$.
\end{definition}

\section{Theoretical Analysis}
\label{sec:theory}
In this section, we explore the global optimal solutions of the $L$-extended unconstrained feature model under the imbalanced data. In the following, we first focus unconstrained feature model on the bias-free case and the bias case, which are followed by the theoretical analysis of $L$-extended unconstrained feature model under the imbalanced data.
\subsection{Unconstrained Feature Model}
We first consider the bias-free case, which corresponds to the following optimization model
\begin{equation}
\label{eq:model-biasfree}
    \min_{W,H} f(W,H)=\frac1{2N}\|WH-Y\|^2_F+\frac{\lambda_W}{2}\|W\|^2_F+\frac{\lambda_H}2\|H\|^2_F.
\end{equation}
Next, we characterize the property of the global optimal solutions of Model \eqref{eq:model-biasfree}, which is stated in Theorem \ref{theo:ufm-biasfree}.
\begin{theo}
\label{theo:ufm-biasfree}
    We consider a dataset with $K$ classes, and there are $n_k$ samples in the $k$-th class for $k=1,\cdots,K$. Define $D=\mathrm{diag}([\sqrt{n_1},\cdots,\sqrt{n_K}])$. 
    Let $W=\begin{bmatrix}
        W_1&\cdots& W_K
    \end{bmatrix}^\top\in\R^{K\times d}$ with $W_k\in\R^d$, $H=\begin{bmatrix}
        H_1&\cdots&H_K
    \end{bmatrix}\in\R^{d\times N}$ with $H_k\in\R^{d\times n_k}$, and $Y=\begin{bmatrix}
        Y_1&\cdots&Y_K
    \end{bmatrix}$ with $Y_k=\begin{bmatrix}
        e_k&\cdots&e_{k}
    \end{bmatrix}\in\R^{K\times n_k}$, where $e_k$ is the $k$-th one-hot vector. Then any global minimizer $(W^*,H^*)$ of \eqref{eq:model-biasfree} satisfies 
    \begin{itemize}
        \item $h_{k,1}=\cdots=h_{k,n_k},\ k=1,\cdots,K$.
        \item  The relationships between $W$ and $H$ are
    \begin{equation*}
    \begin{split}
    W^*\bar H^*=\Sigma^2_*(\Sigma^2_*+N\lambda_WI)^{-1} , W^*(W^*)^\top=D^2\Sigma^2_*(\Sigma^2_*+N\lambda_WI)^{-2},\ 
    (\bar H^*)^\top\bar H^*=\Sigma^2_*D^{-2},
    \end{split}
    \end{equation*}
    where \[\sigma^*_i=\mathop{\arg\min}_{\sigma_i}\frac{n_i\lambda_W}{2(\sigma^2_i+N\lambda_W)}+\frac{\lambda_H}2(\sigma^2_i+N\lambda_W)=\left\{
    \begin{array}{ll}
    \sqrt{\sqrt{\frac{n_i\lambda_W}{\lambda_H}}-N\lambda_W},&\lambda_H\lambda_W\le\frac{n_i}{N^2}\\
    0,&\hbox{otherwise}
    \end{array}
    \right.,\]
    and $\Sigma_*=\mathrm{diag}([\sigma^*_1,\cdots,\sigma^*_K])$.
    \end{itemize}
\end{theo}
\Proof
We provide the proof in another angle, which is deferred to Subsection \ref{proof:theorem_ufm-biasfree}.
\eproof

The proof of Theorem \ref{theo:ufm-biasfree} is not hard. We try to estimate the optimal lower bound of the objective function based on the centroid of each class, which is utilized for minimizer solving. Interestingly, the individual features of each class show variability collapse. Moreover, the $i$-th row of $W$ exhibit orthogonality when $n_i>N^2\lambda_H\lambda_W$; otherwise, they collapse to zero, resulting in an orthogonal geometry. A similar result holds for $W$ and $H$: when 
$n_i>N^2\lambda_H\lambda_W$, the $i$-th row of $W$ and the $i$-th column of $H$ are orthogonal; otherwise, they collapse to zero simultaneously.

Next we focus on the bias case, whose optimization model is
\begin{equation}
\label{eq:model-bias}
    \min_{W,H,b} f(W,H,b)=\frac1{2N}\|WH+b-Y\|^2_F+\frac{\lambda_W}{2}\|W\|^2_F+\frac{\lambda_H}2\|H\|^2_F.
\end{equation}
In the following, we characterize the property of the global optimal solutions of Model \eqref{eq:model-bias}, which is stated in Theorem \ref{theo:ufm-bias}.
\begin{theo}
\label{theo:ufm-bias}
    Let $\vn=[n_1\ \cdots\ N_K]^\top$, $\tilde y_k=e_k-\frac{\vn}{N}$, $\bar h_k=\frac1{n_k}\sum^{n_k}_{i=1}h_{k,i}$, $\bar H=\begin{bmatrix}
    \bar h_1&\cdots&\bar h_K
\end{bmatrix}$ and $\tilde Y=\begin{bmatrix}
    \tilde y_1&\cdots&\tilde y_K
\end{bmatrix}=I_K-1/N\vn\mathbbm{1}^\top_K$. Define $D=\diag([\sqrt{n_1}\ \cdots\ \sqrt{n_K}])$.
Assume $\kappa_1,\cdots,\kappa_K$ are the $K$ eigenvalues of $\tilde YD$, $\kappa=\mathrm{diag}([\kappa_1\ \cdots\ \kappa_K])$ and $\tilde YD=\tilde U\kappa \tilde V^\top$.
Then any global minimizer $(W^*,H^*,b^*)$ of \eqref{eq:model-bias} satisfies  $b^*=\frac{\vn}{N}$,  $h^*_{k,1}=\cdots=h^*_{k,n_k},\ \forall k=1,\cdots,K$, $H^*=[\bar h^*_1 \mathbbm{1}_{n_1}^\top\ \cdots\ \bar h^*_K
 \mathbbm{1}_{n_K}^\top]$ and \begin{equation*}
\begin{split}
\tilde U^\top W^*\bar H^*D\tilde V=&\left(\kappa-N\sqrt{\lambda_W\lambda_H}\right)_+ ,\\
\tilde U^\top W^*(\tilde U^\top W^*)^\top=&\sqrt{\frac{\lambda_H}{\lambda_W}}\left(\kappa-N\sqrt{\lambda_W\lambda_H}I\right)_+,\\
(\bar H^*D\tilde V)^\top\bar H^*D\tilde V=&\sqrt{\frac{\lambda_W}{\lambda_H}}\left(\kappa-N\sqrt{\lambda_W\lambda_H}I\right)_+.
\end{split}
\end{equation*}
\end{theo}
\Proof
We defer the proof to Subsection \ref{proof:theo-ufm-bias}.
\eproof

In the proof, we first provide an estimation of the optimal $b^*$, which implies the global centroid of features is 0. Then we will substitute the optimal value of $b^*$ into the objective function, which is a new bias-free optimization problem, but with a much more complicated label matrix $\hat Y=(I_K-1/N\vn\mathbbm{1}^\top_K)D$.  We based on the singular value decomposition of $\hat Y$ and provide the geometric structure of the optimal minimizer of the UFM under the bias assumption, as stated in Theorem \ref{theo:ufm-bias}. Similarly, the individual features of each class show variability collapse. For the geometric structure, the orthogonal geometry appears on the rows of the left linear transformation of W by the transpose of the left singular matrix and the columns of the right linear transformation of $\bar H$ by the multiplication of $D$ and the right singular matrix.
\subsection{\texorpdfstring{$L$}{}-Extended Unconstrained Feature Model}
We consider the following optimization objective function:
\begin{equation}\label{eq:prob_multilayer}
\begin{split}
    &\min_{W_L,\cdots,W_1,H,b}\ f(W_L,\cdots,W_1,H,b)\\
    =&\min_{W_L,\cdots,W_1,H,b}\ \frac1{2N}\|W_L\cdots W_1H+b\mathbbm{1}^{\top}_N-Y\|^2_F+\frac{1}{2}\sum^L_{j=1}\lambda_{W_j}\|W_j\|^2_F+\frac{\lambda_{H}}2\|H\|^2_F.
\end{split}
\end{equation}
\begin{theo}\label{theo:bias_mult-layer}
Let $W_L\in\R^{K\times d_{L-1}},\cdots,W_1\in\R^{d_1\times d_0},H\in\R^{d_0\times N}$ be a global minimizer of 
the optimization problem \eqref{eq:prob_multilayer}. Let $\hat Y=(I_K-1/N\vn\mathbbm{1}^\top_K)D$ and the corresponding SVD be $\hat Y=\hat U\hat \kappa\hat V^\top$, $r=\min(K,d_{L-1},\cdots,d_0)$. We have the following results:
\begin{itemize}
    \item $h_{k,1}=\cdots=h_{k,n_k},\ k=1,\cdots,K$.
    \item Set $c=\frac{\lambda^{L-1}_{W_1}}{\lambda_{W_L}\lambda_{W_{L-1}}\cdots\lambda_{W_2}}$, $r=\min(d_L,\cdots,d_1,K)$ and $\alpha=\frac{N}{\kappa_k^2}\sqrt[L]{N\lambda_{W_L}\cdots\lambda_{W_1}\lambda_H}$, $x_k^*$ is the largest solution of the equation $\alpha-\frac{x^{L-1}}{(x^L+1)^2}=0$. Define 
    \[\sigma^2_k=\left\{
    \begin{array}{cl}
     \sqrt[L]{\frac{N\lambda_H}{c}}x_k^*,     & \sqrt[L]{N\lambda_{W_L}\cdots\lambda_{W_1}\lambda_H}L^2N\le\kappa^2_k(L-1)^{\frac{L-1}{L}}, \\
     0    & \hbox{otherwise}.
    \end{array}  
    \right..\] 
    Then we have the following 
    \begin{equation*}
        \begin{split}
            &\hat U^\top W_LW_L^\top \hat U=\frac{\lambda_{W_1}}{\lambda_{W_M}}\Upsilon_1,\ \hat U^\top W_L\cdots W_1W^\top_1\cdots W_L^\top\hat U=c\Upsilon^L_1,\\
            &\hat V^\top D\bar H^\top\bar HD\hat V=c\Upsilon_2\kappa^2,\ \hat U^\top W_M\cdots W_1\bar H D\hat V=c\Upsilon_2\kappa,
        \end{split}
    \end{equation*}
        where 
        \begin{align*}
            \Upsilon_1=&\begin{bmatrix}
                \diag(\begin{bmatrix}
                \sigma^2_1 & \cdots &\sigma^2_r
            \end{bmatrix})&0_{r, K-r}\\
            0_{K-r,r}&0_{K-r,K-r}
            \end{bmatrix},\\
            \Upsilon_2=&\begin{bmatrix}
                \diag\left(\begin{bmatrix}
                \frac{\sigma^{2L}_1}{(c\sigma^{2L}_1+N\lambda_H)^2} & \cdots &\frac{\sigma^{2L}_r}{(c\sigma^{2L}_r+N\lambda_H)^2}
            \end{bmatrix}\right)&0_{r, K-r}\\
            0_{K-r,r}&0_{K-r,K-r}
            \end{bmatrix}.
        \end{align*}
    \item Let $$W=\hat U^\top W_L\cdots W_1=\begin{bmatrix}
        w_1^\top\\\vdots\\w_K^\top
    \end{bmatrix},\ G=\bar H D\hat V=\begin{bmatrix}
         g_1 &\cdots &g_K
    \end{bmatrix}.$$ Then 
    $$
    \left\{
    \begin{array}{ll}
       \frac{w_i}{\|w_i\|}=\frac{g_i}{\|g_i\|},  & \sqrt[L]{N\lambda_{W_L}\cdots\lambda_{W_1}\lambda_H}L^2N\le\kappa^2_k(L-1)^{\frac{L-1}{L}}, \\
       w_i=g_i=0,  & \rm{otherwise}.
    \end{array}
    \right.
    $$
\end{itemize}

\end{theo}
\Proof
We defer the proof to Subsection \ref{proof:bias-multilayer}.
\eproof

Theorem \ref{theo:bias_mult-layer} is an extended result of Theorem \ref{theo:ufm-bias}. Similar as the proof of Theorem \ref{theo:ufm-bias}, we also compute the optimal value of $b^*$ first and then substitute it into the objective function, which will be used to get the global optimizer $(W^*_L,\cdots,W^*_1,\bar H^*)$. Moreover, the rows of $\hat U^\top W^*_L\cdots W^*_1$  and the columns of $\bar HD\hat V$ collapse if the sign of $\sqrt[L]{N\lambda_{W_L}\cdots\lambda_{W_1}\lambda_H}L^2N-\kappa^2_k(L-1)^{\frac{L-1}{L}}$ is negative, otherwise the correspondence of them are orthogonal.
\subsection{The Estimation of Singular Values}
The results of Theorems \ref{theo:ufm-bias} and \ref{theo:bias_mult-layer} are based on the singular values of $\hat Y=(I_K-1/N\vn\mathbbm{1}^\top_K)D$. In the following, we give the estimation of the singular values, which is stated in Theorem \ref{theo:singular}. 
We consider an imbalanced dataset according to the number of samples in each class, denoted as $(\{N_i,\ell_i\}^m_{i=1},\{n_j\}^K_{j=1})$-Imbalances, which is defined as follows:
\begin{definition}[$(\{N_i,\ell_i\}^m_{i=1},\{n_j\}^K_{j=1})$-Imbalances]
\label{def:imbalance}
    We consider a dataset with $K$ classes, where there are $n_k$ samples in the $k$-th class for $k=1,\ldots,K$. We assume there are $m$ groups in the dataset based on the number of samples in each class, with $\ell_i$ representing the number of classes that contain $N_i$ samples for $i=1,\ldots,m$. Let $\{n_{[i]}\}_{i=1}^K$ be the rearrangement of $\{n_{[i]}\}_{i=1}^K$ in non-increasing order. We define the dataset as $(\{N_i,\ell_i\}_{i=1}^m,\{n_j\}_{j=1}^K)$-imbalanced if 
    $$n_{[\sum^{j-1}_{i=1}\ell_i+1]}=\cdots=n_{[\sum^{j}_{i=1}\ell_i]}=N_j,j=1,\cdots,m.$$ 
\end{definition}
From the above definition, it is obvious that  $N_1>N_2>\cdots>N_m$.

\begin{figure}[ht]
	\begin{center}
		\subfigure{
			\includegraphics[width=4.5cm, height=4cm]{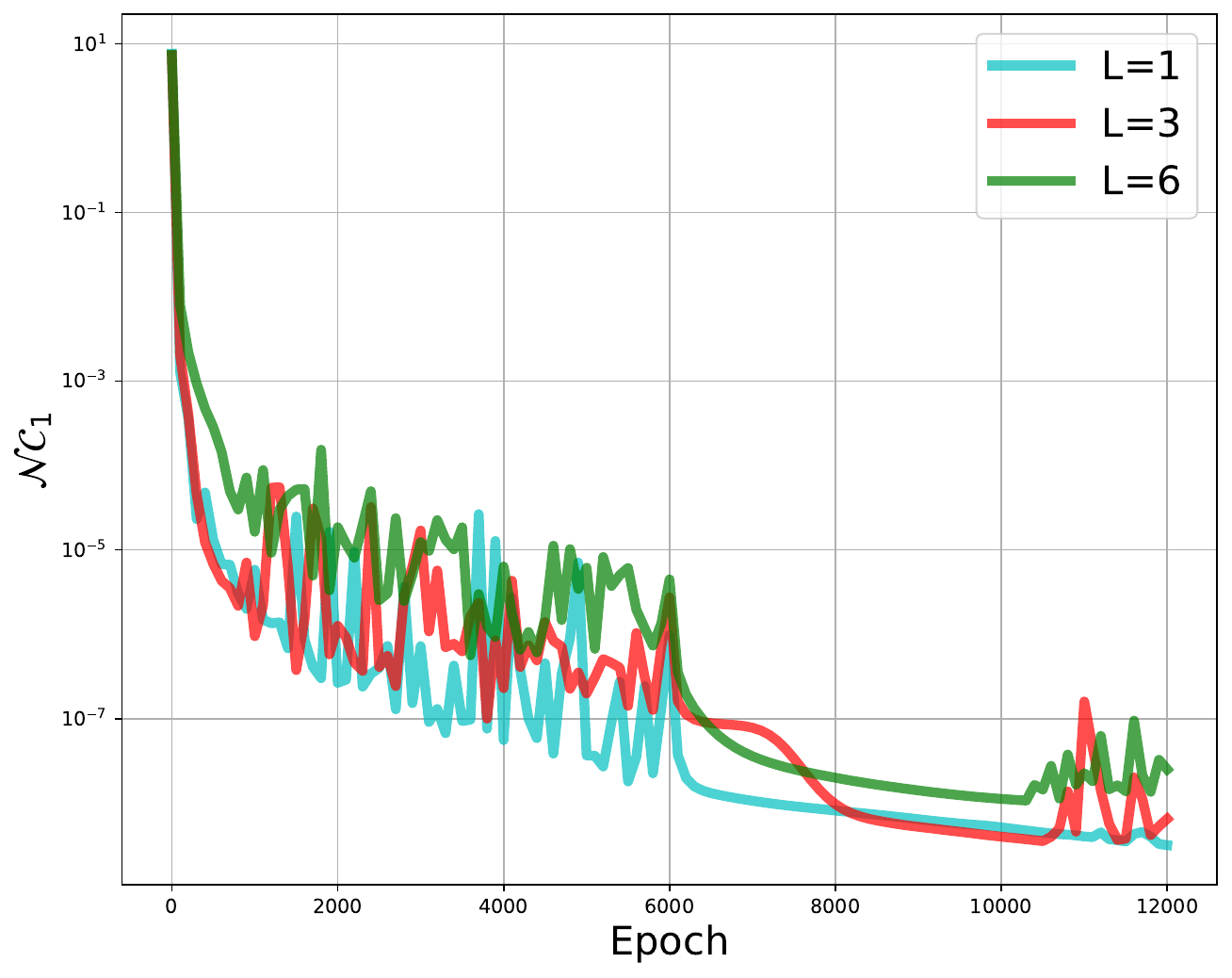}}
		\subfigure{
			\includegraphics[width=4.5cm, height=4cm]{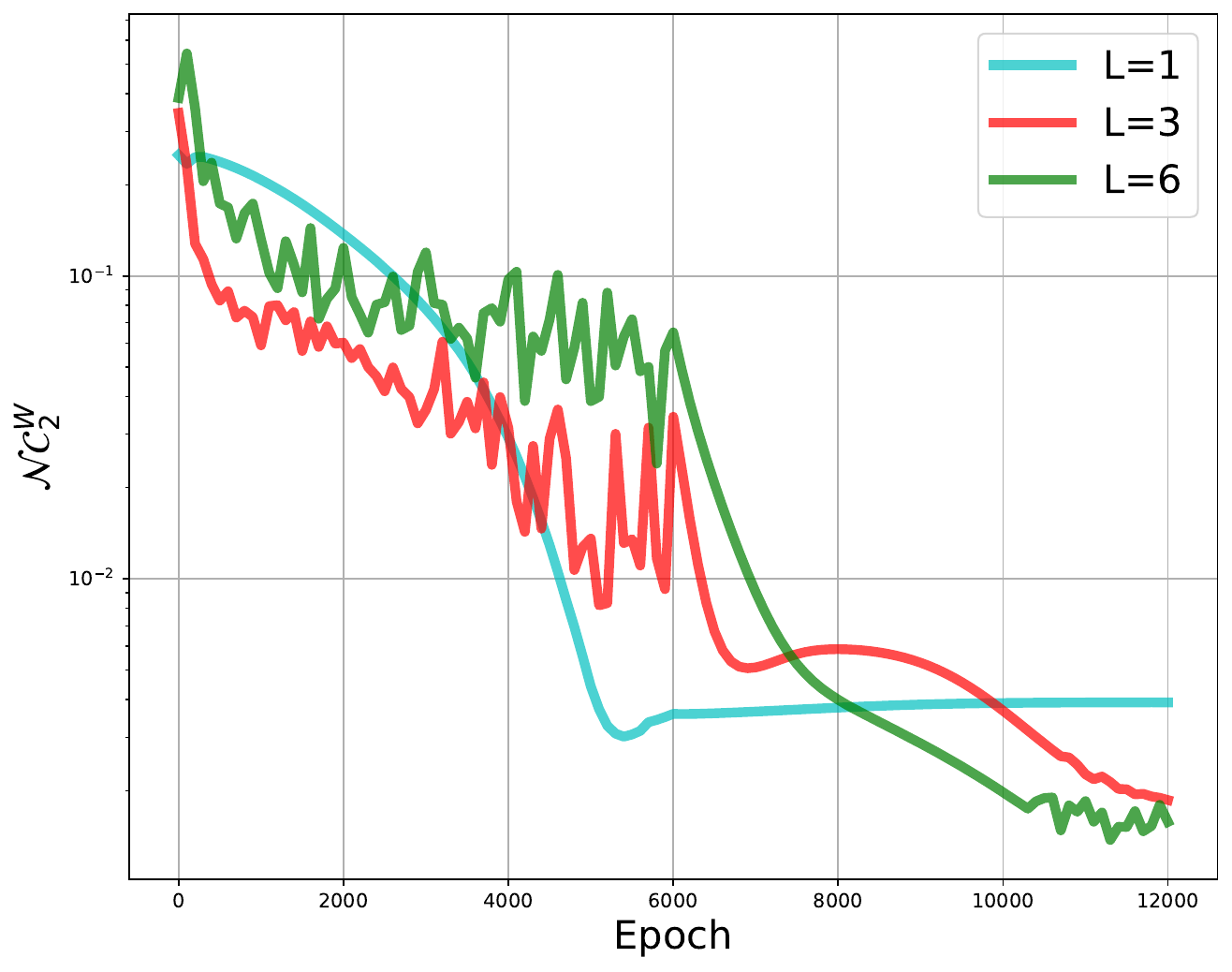}}
        \subfigure{
			\includegraphics[width=4.5cm, height=4cm]{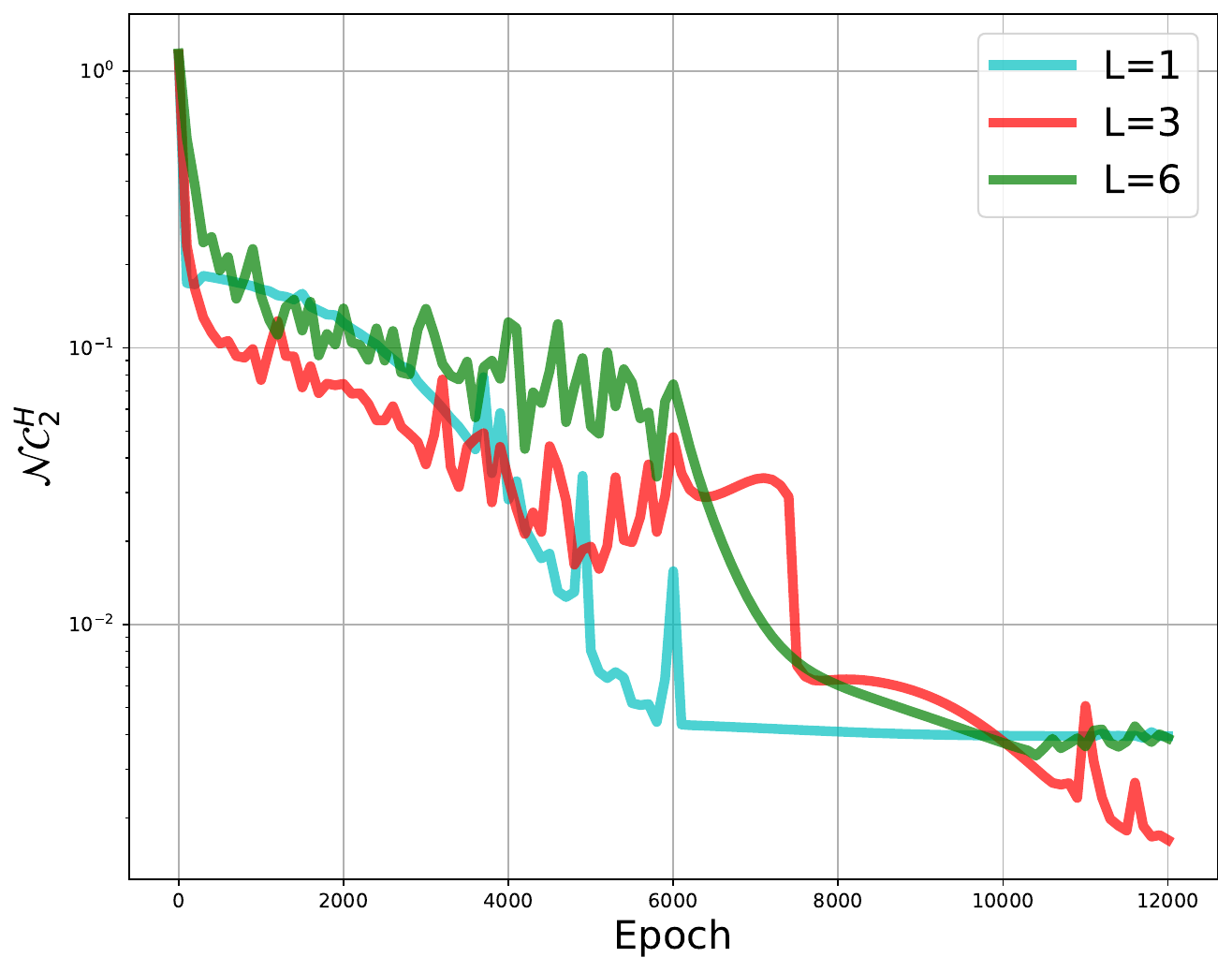}}
        \subfigure{
			\includegraphics[width=4.5cm, height=4cm]{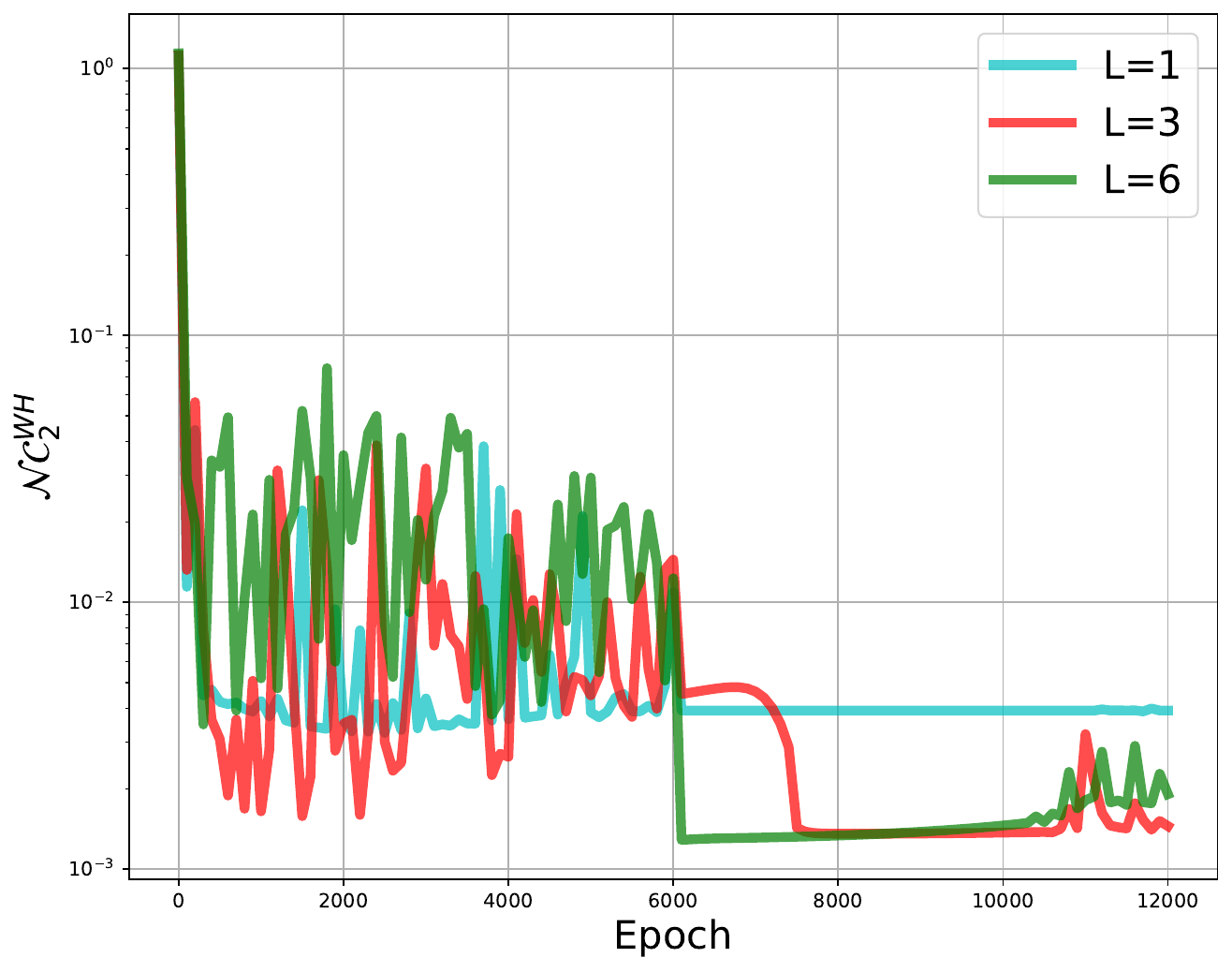}}
        \subfigure{
			\includegraphics[width=4.5cm, height=4cm]{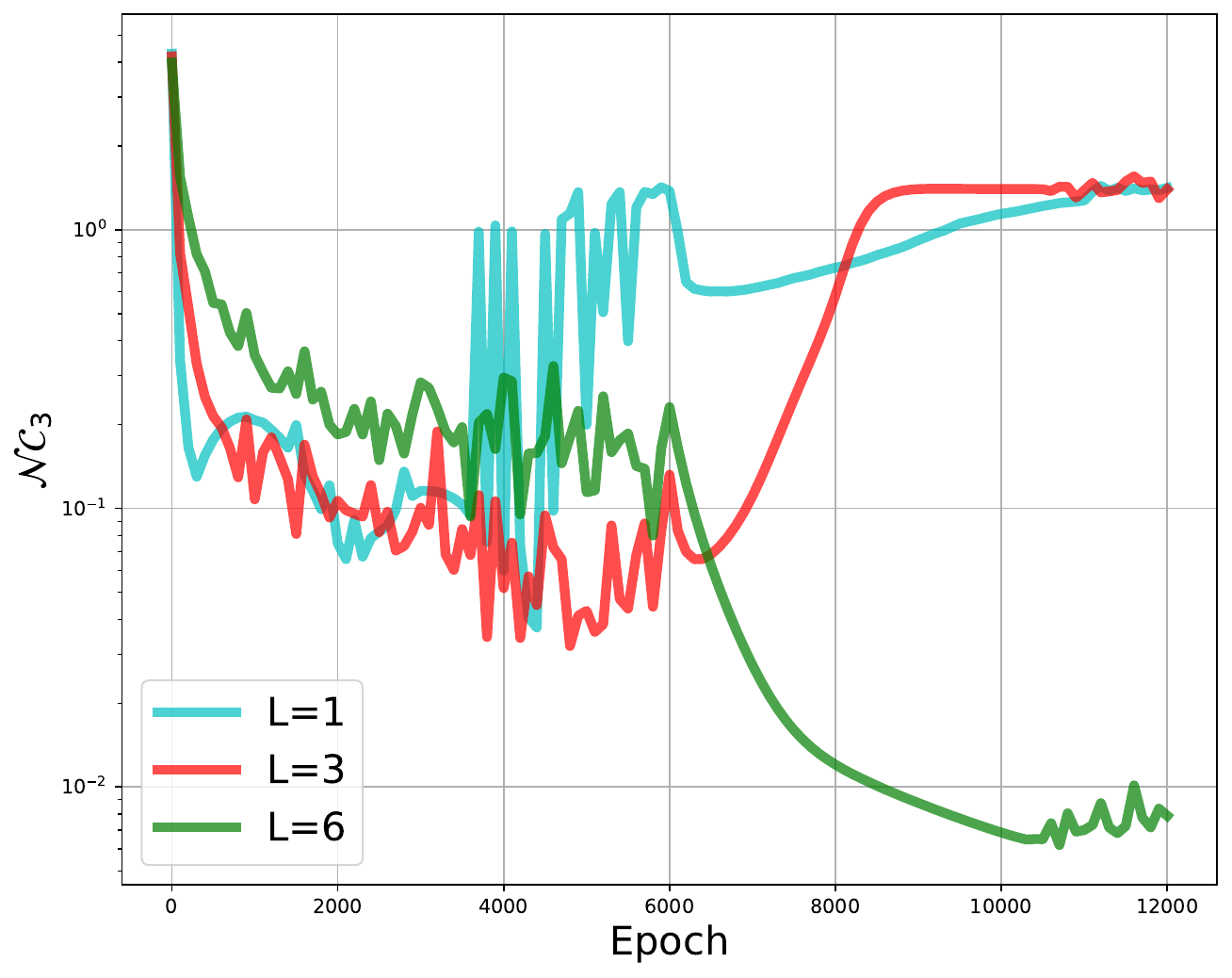}}
        \subfigure{
			\includegraphics[width=4.5cm, height=4cm]{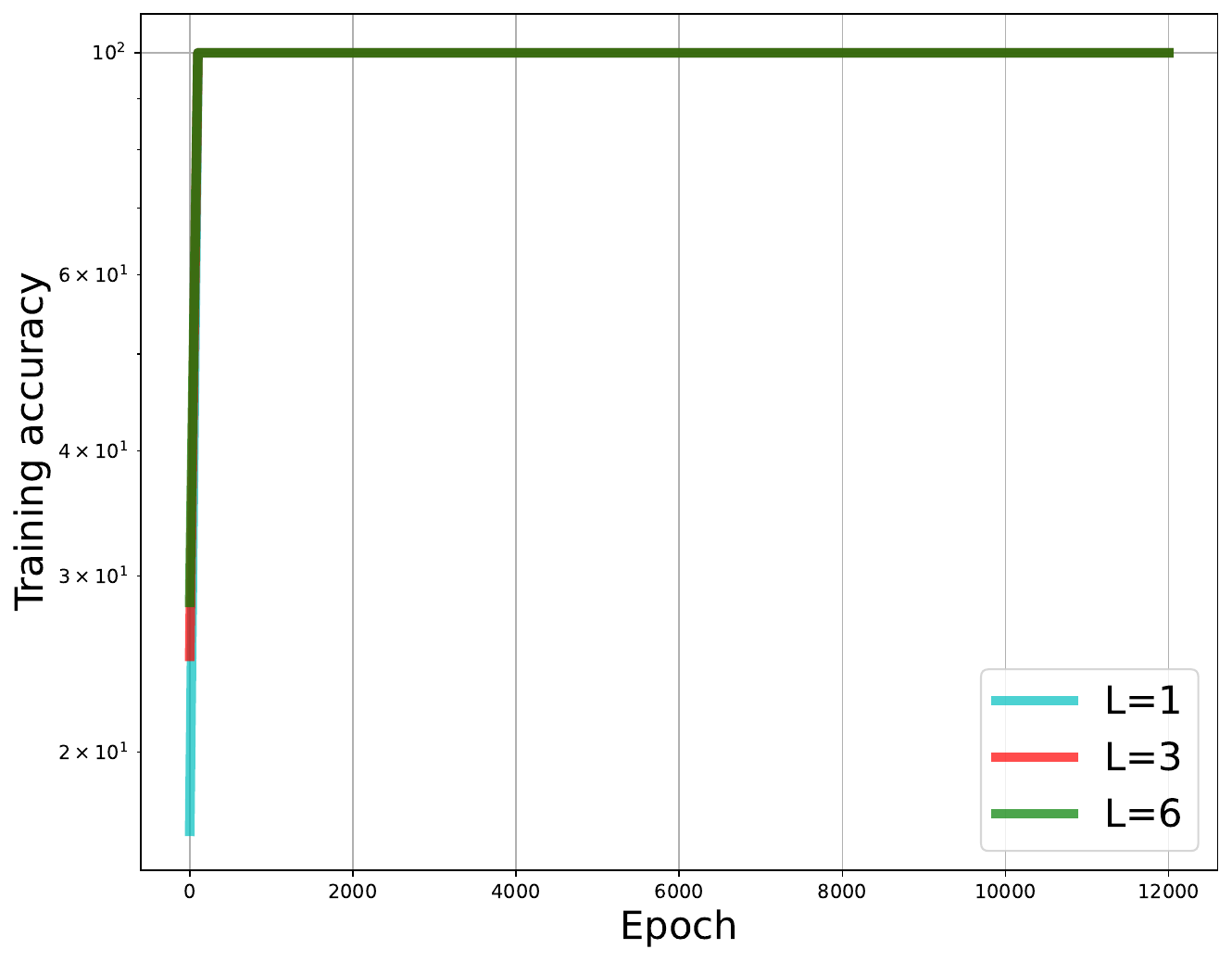}}
		\caption{Bias Case: The performance of the NC metrics and training accuracy versus epoch with a 6-layer MLP backbone on an imbalanced subset of CIFAR10 for $L$-extended unconstrained feature model with $L=1,3,6$.} \label{fig:mlp-cifar}
	\end{center}
\end{figure}
\begin{theo}
\label{theo:singular}
    We consider a dataset is with $(\{N_i,\ell_i\}^m_{i=1},\{n_j\}^K_{j=1})$-Imbalances  defined in Definition \ref{def:imbalance}. Without loss of generality we assume $\vn=\begin{bmatrix}
        n_1&\cdots&n_K
    \end{bmatrix}^\top=\begin{bmatrix}
    N_1\mathbbm{1}^\top_{\ell_1}&\cdots&N_m\mathbbm{1}^\top_{\ell_m}
    \end{bmatrix}^\top$ and $\sqrt{\vn}=\begin{bmatrix}
        \sqrt{N_1}\mathbbm{1}^\top_{\ell_1}&\cdots&\sqrt{N_m}\mathbbm{1}^\top_{\ell_m}
    \end{bmatrix}^\top$. Let $ D=\mathrm{diag}(\sqrt{\vn})$, $N=\sum^K_{j=1}n_j$ and $\tilde Y=(I_K-1/N\vn\mathbbm{1}^\top_K)D=D-\frac{\vn}{N}(\sqrt{\vn})^\top$. Set $$G=\begin{bmatrix}
    \sqrt{N_1}(1-\frac{N_1\ell_1}{N})&-\frac{N_1\sqrt{N_2}}{N}\sqrt{\ell_1\ell_2}&\cdots&-\frac{N_1\sqrt{N_m}}{N}\sqrt{\ell_1\ell_m}\\
    -\frac{N_2\sqrt{N_1}}{N}\sqrt{\ell_2\ell_1}&\sqrt{N_2}(1-\frac{N_2\ell_2}{N})&\cdots&-\frac{N_2\sqrt{N_1}}{N}\sqrt{\ell_2\ell_m}\\
    \vdots&\vdots&\ddots&\vdots\\
    -\frac{N_m\sqrt{N_1}}{N}\sqrt{\ell_m\ell_1}&-\frac{N_m\sqrt{N_2}}{N}\sqrt{\ell_m\ell_2}&\cdots&\sqrt{N_m}(1-\frac{N_m\ell_m}{N})\\
\end{bmatrix},$$ and $\tilde\sigma_j,j=1,\cdots,m$ are the singular values of $G$ satisfying $\tilde\sigma_1\ge\cdots\ge\tilde\sigma_m$. Then the singular values of $\tilde Y$ satisfy 
\[s_i=\left\{
\begin{array}{ll}
    \sqrt{N_1}, & i=1,\cdots,\ell_1-1, \\
    \sqrt{N_j}, & i=\sum^j_{\alpha=1}\ell_{\alpha-1}+1,\cdots,\sum^j_{\alpha=1}\ell_\alpha-1,\ j=2,\cdots,m,\\
    \tilde\sigma_i, & i=\sum^j_{\alpha=1}\ell_\alpha,\ j=1,\cdots,m,
\end{array}
\right.\]
and $0=\tilde\sigma_m<\sqrt{N_m}<\tilde\sigma_{m-1}<\sqrt{N_{m-1}}<\cdots<\sqrt{N_2}<\tilde\sigma_1<\sqrt{N_1}$. 

Especially, 
\begin{itemize}
        \item when $m=2$, the singular values of $G$ are $0$ and $\sqrt{\frac{KN_1N_2}{N}}$.
        \item when $m=3$, the singular values of $G$ are $0$ and the roots of quadratic polynomial $\lambda^2-a\lambda+b=0$, where 
        $a = \frac 1N[N_1(N_2\ell_2+N_3\ell_3)+n_2(N_1\ell_1+N_3\ell_3)+N_3(N_1\ell_1+N_2\ell_2)]$, and $b = \frac{KN_1N_2N_3}{N}$.
    \end{itemize}
\end{theo}
\Proof
We will defer the proof to Subsection \ref{subsec:proof_singular}. The theorem is divided into two parts: Theorem \ref{theo:singular_proof} presents the singular values of $\hat Y$ in relation to the singular values of $G$. The proof of the singular values of $G$ is provided in Theorem \ref{theo:singular_proof1} for $m=2,3$. 
\eproof
\section{Numerical Results}\label{sec:numerical}
In this section, we present numerical experiments to validate our theoretical results concerning imbalanced data. We utilize a 6-layer MLP as the backbone feature extractor to generate the 'unconstrained' features, followed by a linear network with 
$L=1,3,6$ layers. The hidden width $d=2048$ is chosen for both the feature extraction and the deep linear model. We consider the objective function \eqref{eq:prob_multilayer} and observe the NC phenomenon. To evaluate performance, we employ the following three NC metrics:
\begin{itemize}
    \item \textbf{Feature collapse:} let $h_{k,i}$ be the features of the $i$-th sample in the $k$-th class. Then the class-means and global mean are
    \[h_k=\frac{1}{n_k}\sum^{n_k}_{i=1}h_{k,i},\ h_G=\frac{1}{\sum^K_{k=1}n_k}\sum^K_{k=1}\sum^{n_k}_{i=1}h_{k,i},\]
    and the within-class, between-class covariance matrices are
    \[\Sigma_W=\frac{1}{\sum^K_{k=1}n_k}\sum^K_{k=1}\sum^{n_k}_{i=1}(h_{k,i}-h_k)(h_{k,i}-h_k)^\top,\ \Sigma_B=\frac{1}{K}\sum^K_{k=1}(h_{k}-h_G)(h_{k}-h_G)^\top.\]
    Then the NC1 metric is 
    \[NC_1=\frac{1}{K}\tr(\Sigma_W\Sigma^{\dagger}_B).\]
    \item \textbf{Convergence to Geometric Structure:} Define $\bar H=[h_1 \cdots h_K]$, $\vn=[n_1 \cdots n_K]^\top$, $D=\diag([\sqrt{n_1} \cdots \sqrt{n_K}])$ and $\hat U\kappa\hat V^\top=(I_K-\frac{1}{N}\vn\mathbbm{1}^\top_K)D$. Then the NC2 and NC3 metrics are
    \begin{align*}
        NC^W_2=&\left\|\frac{\hat U^\top(W_M\cdots W_1)(W_M\cdots W_1)^\top\hat U}{\|(W_M\cdots W_1)(W_M\cdots W_1)^\top\|_F}-\frac{\Upsilon^L_1}{\|\Upsilon^L_1
    \|_F}\right\|_F,\\
        NC^{H}_2=&\left\|\frac{\hat V^\top D\bar H^\top\bar HD\hat V}{\|\hat V^\top D\bar H^\top\bar HD\|_F}-\frac{\Upsilon_2\kappa^2}{\|\Upsilon_2\kappa^2
        \|_F}\right\|_F,\\
        NC^{WH}_2=&\left\|\frac{\hat U^\top W_M\cdots W_1\bar HD\hat V}{\|W_M\cdots W_1\bar HD\|_F}-\frac{\Upsilon_2\kappa}{\|\Upsilon_2\kappa
    \|_F}\right\|_F,
    \end{align*}
    where $\Upsilon_1,\Upsilon_2$ are defined in Theorem \ref{theo:bias_mult-layer}.\\
    Let $W=\hat U^\top W_L\cdots W_1$ and $W^\top=\begin{bmatrix}
        w_1&\cdots&w_K
    \end{bmatrix}$, $G=\bar HD\hat V=\begin{bmatrix}
        g_1&\cdots & g_K
    \end{bmatrix}$, then
    \[NC_3=\left\|\begin{bmatrix}
        \frac{w_1}{\|w_1\|_2}&\cdots&\frac{w_K}{\|w_k\|_2}
    \end{bmatrix}-\begin{bmatrix}
        \frac{g_1}{\|g_1\|_2}&\cdots&\frac{g_K}{\|g_K\|_2}
    \end{bmatrix}\right\|_F.\]
\end{itemize}
Our focus is on two specific datasets: CIFAR10 and EMNIST letter. We choose a random subset of CIFAR10 dataset with number of training samples of each class from the list \{500, 500, 400, 400, 300, 300, 200, 200, 100, 100\}. For EMNIST letter dataset, we randomly sample 1 major class with 1500 samples, 5 medium classes with 600 samples per class and 20 minor classes with 50 samples per class.

\begin{figure}[ht]
	\begin{center}
		\subfigure{
			\includegraphics[width=4.5cm, height=4cm]{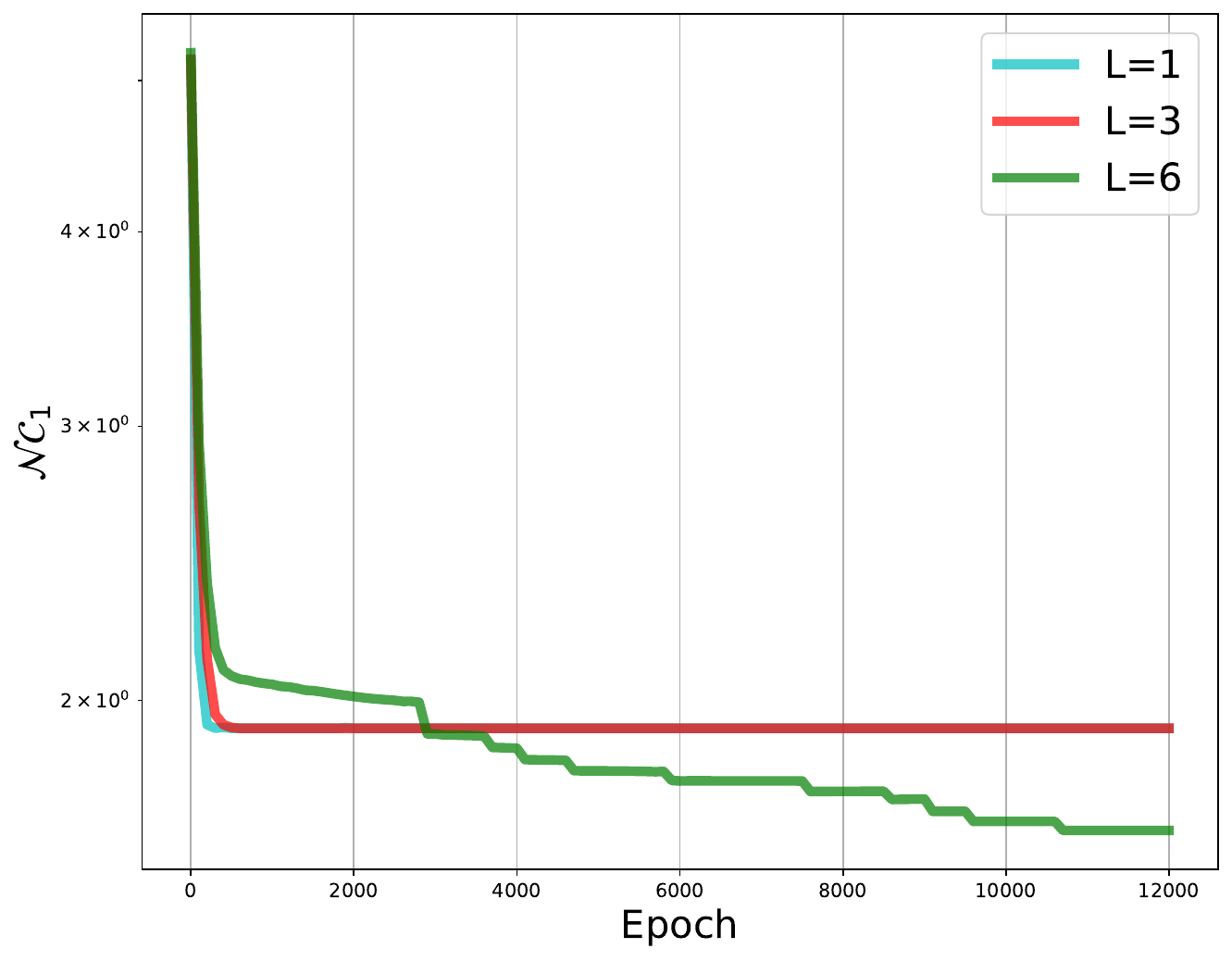}}
		\subfigure{
			\includegraphics[width=4.5cm, height=4cm]{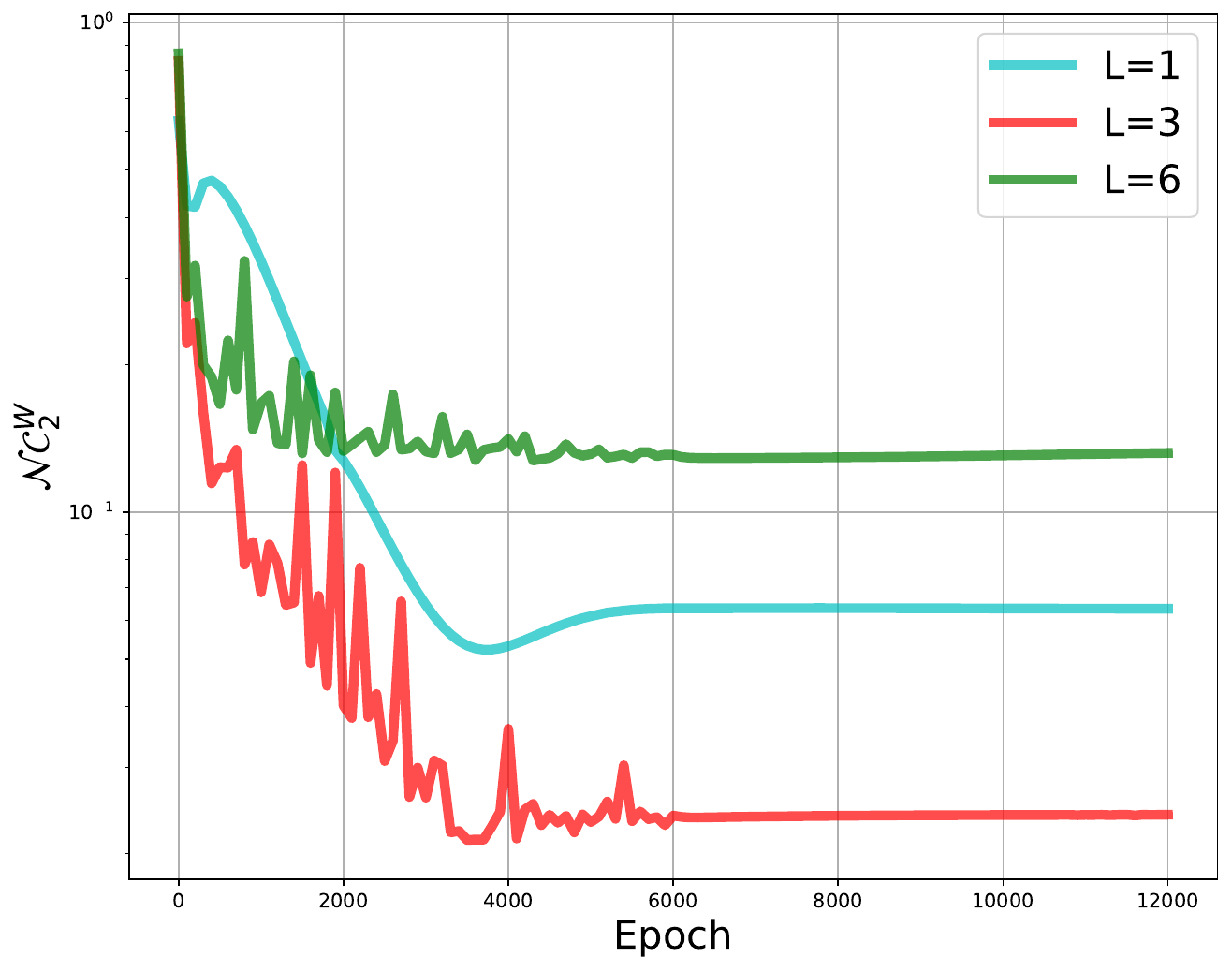}}
        \subfigure{
			\includegraphics[width=4.5cm, height=4cm]{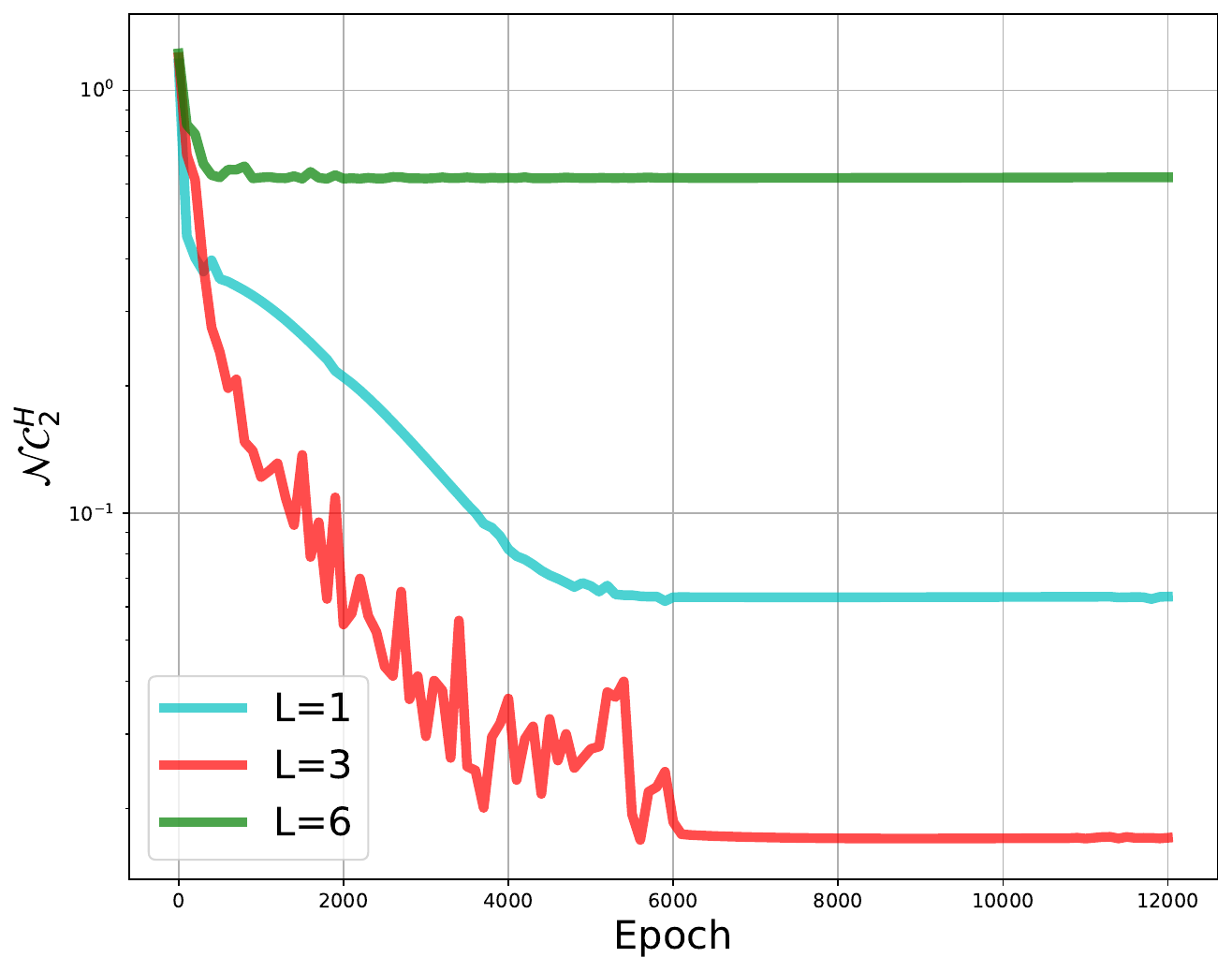}}
        \subfigure{
			\includegraphics[width=4.5cm, height=4cm]{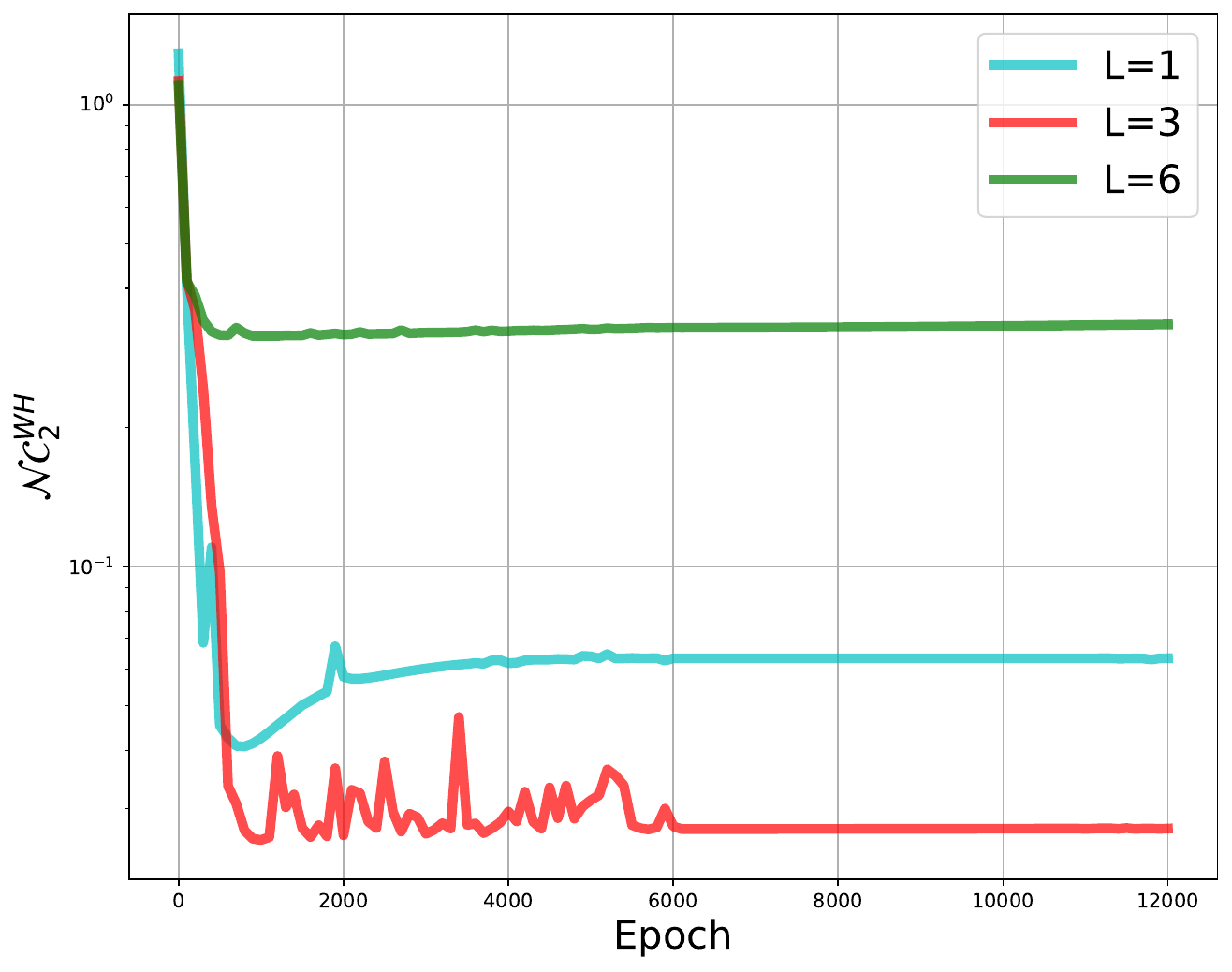}}
        \subfigure{
			\includegraphics[width=4.5cm, height=4cm]{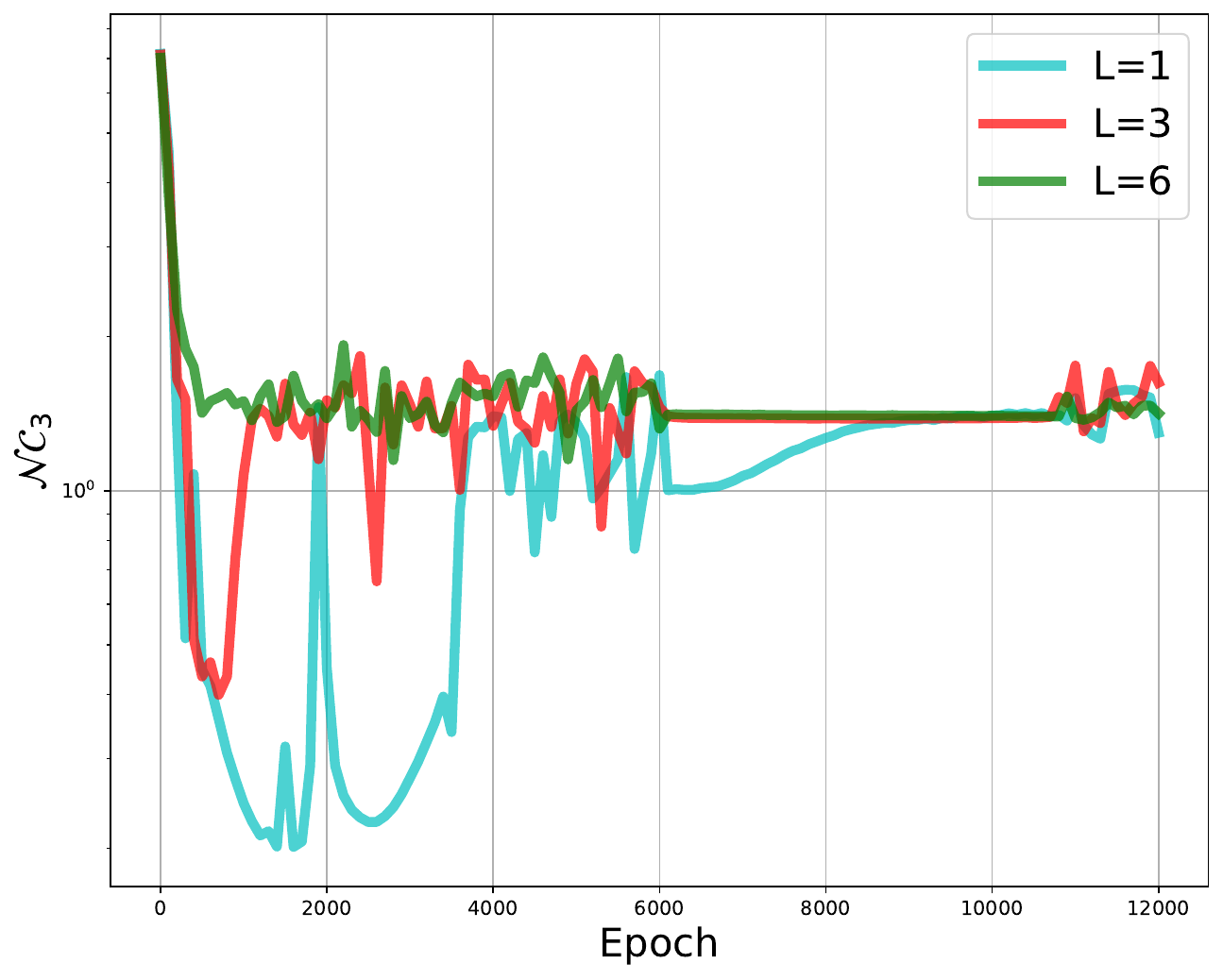}}
        \subfigure{
			\includegraphics[width=4.5cm, height=4cm]{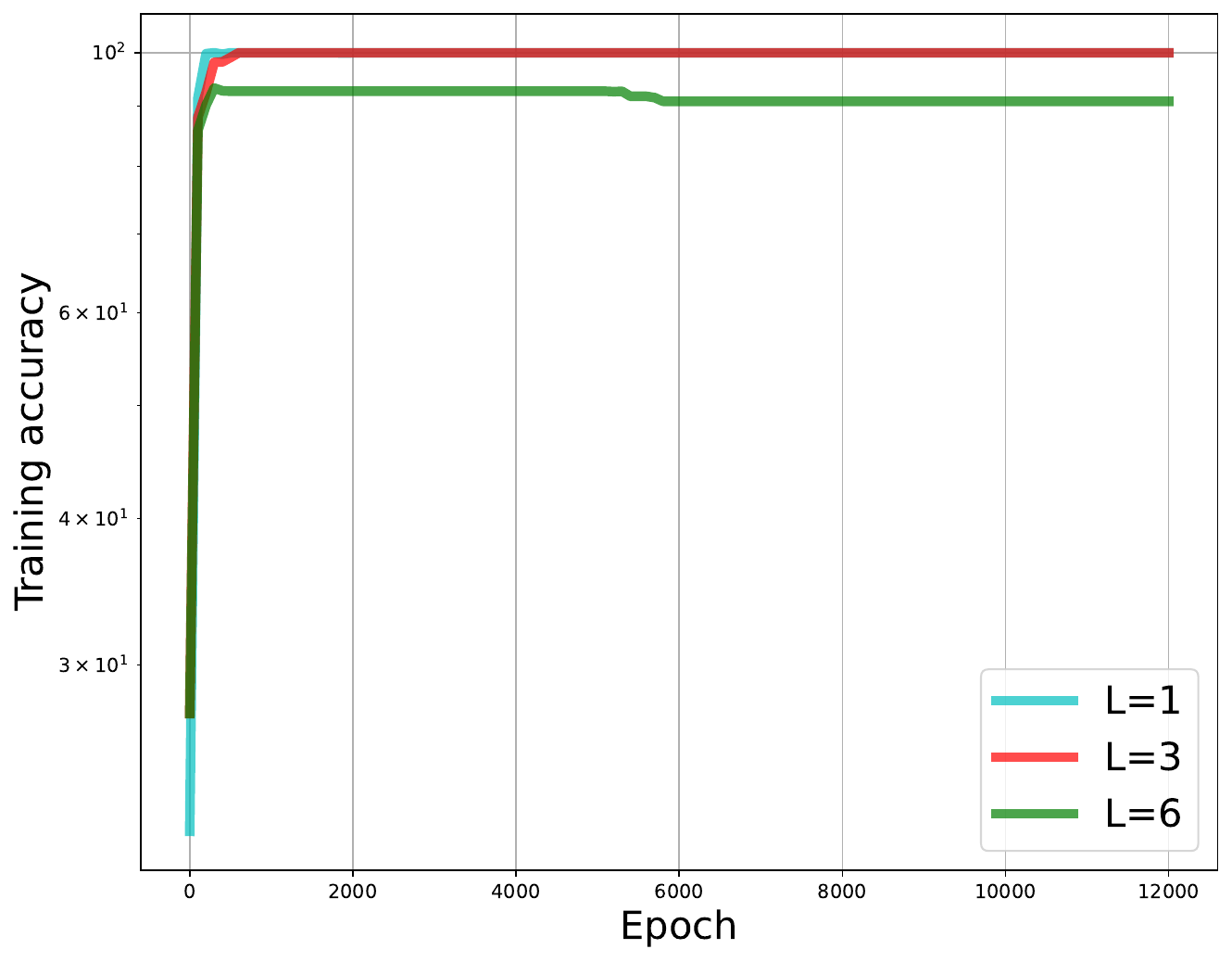}}
		\caption{Base case: The performance of the NC metrics and training accuracy versus epoch with a 6-layer MLP backbone on an imbalanced subset of EMNIST for $L$-extended unconstrained feature model with $L=1,3,6$.} \label{fig:mlp-emnist}
	\end{center}
\end{figure}

We train it using the Adam optimizer with an initial learning rate of 0.0001. The model undergoes training for a total of 12000 epochs and set the batchsize equal to 128. Figures \ref{fig:mlp-cifar} and \ref{fig:mlp-emnist}  provide the quantitative comparison results of the above metrics and training accuracy versus epoch with $L=1,3,6$ and a 6-layer MLP backbone on CIFAR10 and EMNIST datasets. We also try the bias-free cases on the CIFAR10 and EMNIST datasets, the numerical experiments are deferred to \ref{appendix:numerical_bias-free}.

\section{Conclusion}\label{sec:conclusion}
In this paper, we analyze the global optimal solution of the $L$-extended unconstrained feature model with imbalanced data, which corresponds to a mean squared error model that includes biases. When 
$L$=1, it reduces to the unconstrained feature model. Our findings are as follows: (1) The features within the same class converge to their class mean, similar to both the balanced case and the imbalanced case without bias. (2) The rows of the left orthonormal transformation of the linear classifier are orthogonal, and the columns of the right orthonormal transformation of the class-mean matrix are also orthogonal. (3) The $i$-th row of the left orthonormal transformation of the linear classifier aligns with the $i$-th column of the right orthonormal transformation of the class-mean matrix. For $L>1$, we observe similar results, where the linear classifier is replaced by the products of $L$ linear classifiers. Numerical experiments support our theoretical findings.

There are some limitations to this work that will be addressed in future papers: (1) Nonlinear transformations are excluded in the final layers. (2) We do not discuss the impact of the neural collapse phenomenon on downstream tasks.
\bibliographystyle{alpha}
\bibliography{nc}
\newpage
\appendix
\section{Auxiliary Lemma}
\begin{lem}[Lemma 2.3 in \cite{zhu2021geometric}]\label{lem:cite_2.3}
For any fixed $Z\in\mathbb{R}^{K\times N}$ and $\alpha>0$, we have 
\[\|Z\|_*=\min_{Z=WE}\frac12\left(\frac1{\alpha}\|W\|^2_F+\alpha\|E\|^2_F\right),\]
where $\|Z\|$ is the nuclear norm of $Z$. Moreover, the minimizers $W$ and $E$ obey $W=\alpha^{-\frac{1}{4}}U\Sigma^{\frac{1}{2}}R^\top$, $H=\alpha^{\frac{1}{4}}R\Sigma^\frac{1}{2}V^\top$, where $U\Sigma V^\top$ is the SVD of $Z$ and $R$ is an orthonormal matrix with suitable dimension.
\end{lem}
\begin{lem}[Minimizer of the function $g(x)=\frac{1}{ x^L+1}+\alpha x$, D.2.1 in \cite{dang2023neural}]\label{lem:min_g}
We consider the function $g(x)=\frac{1}{ x^L+1}+\alpha x$ with $x\ge 0$, any $\alpha>0$ and $L\ge 2$. The minimizer of the function $g(x)$ is related to the value of $\alpha$:
\begin{itemize}
    \item When $\alpha>\frac{(L-1)^{\frac{L-1}{L}}}{L}$, the minimizer of $g(x)$ is $x=0$,
    \item when $\alpha=\frac{(L-1)^{\frac{L-1}{L}}}{L}$, the minimizer of $g(x)$ is $x=0$ or $x=(L-1)^{1/L}$,
    \item when $\alpha<\frac{(L-1)^{\frac{L-1}{L}}}{L}$, the minimizer of $g(x)$ is the largest solution of the equation $\alpha-\frac{Lx^{L-1}}{(x^L+1)^2}=0$.
\end{itemize}
\end{lem}
\section{Technical Details in Proof}
\label{sec:details}
\subsection{Proof of Theorem \ref{theo:ufm-biasfree}}\label{proof:theorem_ufm-biasfree}
\Proof
Let $H_k=\begin{bmatrix}
    h_{k,1}&\cdots&h_{k,n_k}
\end{bmatrix}\in\R^{d\times n_k}$, $\bar h_k=\frac1{n_k}\sum^{n_k}_{i=1}h_{k,i}$ and $\bar H=\begin{bmatrix}
    \bar h_1 & \cdots & \bar h_K
\end{bmatrix}\in\R^{d\times K}$. Then 
\begin{equation*}
\label{eq:}
\begin{split}
f(W,H)=&\frac1{2N}\|WH-Y\|^2_F+\frac{\lambda_W}{2}\|W\|^2_F+\frac{\lambda_H}2\|H\|^2_F\\
=&\frac1{2N}\sum^K_{k=1}n_k\frac1{n_k}\sum^{n_k}_{i=1}\|Wh_{k,i}-y_k\|^2_2+\frac{\lambda_W}2\|W\|^2_F+\frac{\lambda_H}2\sum^K_{k=1}n_k\frac1{n_k}\sum^{n_k}_{i=1}\|h_{k,i}\|^2_2\\
\ge&\frac1{2N}\sum^K_{k=1}n_k\left\|W\left(\frac1{n_k}\sum^{n_k}_{i=1}h_{k,i}\right)-y_k\right\|^2_2+\frac{\lambda_W}2\|W\|^2_F+\frac{\lambda_H}2\sum^K_{k=1}n_k\left\|\frac1{n_k}\sum^{n_k}_{i=1}h_{k,i}\right\|^2_2\\
=&\frac1{2N}\sum^K_{k=1}n_k\left\|W\bar h_k-y_k\right\|^2_2+\frac{\lambda_W}2\|W\|^2_F+\frac{\lambda_H}2\sum^K_{k=1}n_k\left\|\bar h_k\right\|^2_2
\triangleq \tilde f(W,\bar H),
\end{split}
\end{equation*}
where $f(W,H)=\tilde f(W,\bar H)$ holds if and only if $h_{k,1}=\cdots=h_{k,n_k},\forall k\in[K]$. Define $E=\bar H D$, then
\begin{equation*}
\begin{split}
\tilde f(W,\bar H)
=&\frac1{2N}\|W\bar HD-D\|^2_F+\frac{\lambda_W}{2}\|W\|^2_F+\frac{\lambda_H}2\|\bar HD\|^2_F\\
=&\frac1{2N}\|WE-D\|^2_F+\frac{\lambda_W}{2}\|W\|^2_F+\frac{\lambda_H}2\|E\|^2_F\triangleq \hat f(W,E).
\end{split}
\end{equation*}
Therefore,
\begin{equation}
\label{eq:model_z}
\begin{split}
&\min_{W,E}\ \hat f(W,E)=\min_{W,E}\ \frac1{2N}\|WE-D\|^2_F+\frac{\lambda_W}{2}\|W\|^2_F+\frac{\lambda_H}2\|E\|^2_F\\
=&\min_Z\ \left(\frac1{2N}\|Z-D\|^2_F+\min_{W,E,Z=WE}\frac{\lambda_W}{2}\|W\|^2_F+\frac{\lambda_H}2\|E\|^2_F\right)\\
=&\min_Z\ \frac1{2N}\|Z-D\|^2_F+\sqrt{\lambda_W\lambda_H}\|Z\|_*\triangleq \min_Z\ g(Z).
\end{split}
\end{equation}
Here the second equation is from Lemma \ref{lem:cite_2.3}. Note that $g(Z)$ is convex about $Z$, the minimizer of \eqref{eq:model_z} about $Z$ is $Z^*=(D-N\sqrt{\lambda_W\lambda_H})_+=\mathrm{diag}([(\sqrt{n_1}-N\sqrt{\lambda_W\lambda_H})_+\ \cdots\ \sqrt{n_K}-N\sqrt{\lambda_W\lambda_H})_+])$ according to singular value thresholding algorithm \cite{cai2010singular}. 

Next we analyze the critical point of $\hat f(W,E)$. We compute the derivatives of $\hat f(W,E)$ about $W$ and $E$ respectively and get
\begin{eqnarray}
\frac{\partial \hat f}{\partial W}=\frac1N(WE-D)E^\top+\lambda_WW=\frac1N[W(EE^\top+N\lambda_WI)-DE^\top],\\
\frac{\partial \hat f}{\partial E}=\frac1NW^\top(WE-D)+\lambda_HE=\frac1N[(W^\top W+N\lambda_HI)E-W^\top D].
\end{eqnarray}
Let the singular value decomposition of $E$ be $E=U\Sigma V^\top$. From $\frac{\partial \hat f}{\partial W}=0$, we can get
\begin{equation}
\begin{split}
W=&DE^\top(EE^\top+N\lambda_WI)^{-1}=DV\Sigma U^\top U(\Sigma V^\top V\Sigma +N\lambda_WI)^{-1}U^\top\\
=&DV\Sigma U^\top U(\Sigma^2+N\lambda_WI)^{-1}U^\top=DV\Sigma(\Sigma^2+N\lambda_WI)^{-1}U^\top.
\end{split}
\end{equation}
When $W,E$ arrive at the critical points, $Z^*=WE=DV\Sigma(\Sigma^2+N\lambda_WI)^{-1}U^\top U\Sigma V^\top=DV\Sigma(\Sigma^2+N\lambda_WI)^{-1}\Sigma V^\top$. That is, $D^{-1}Z^*=V\Sigma(\Sigma^2+N\lambda_WI)^{-1}\Sigma V^\top$. Note that $Z^*$ is a diagonal matrix, then $D^{-1}Z^*$ is also a diagonal matrix, then $V=I$. 
Therefore, minimizing $\hat f$ can be reformulate as the following
\begin{eqnarray*}
&&\mathop{\arg\min}_{W,E}\hat f(W,E)\\
&\Leftrightarrow&\mathop{\arg\min}_\Sigma\frac1{2N}\|D\Sigma^2(\Sigma^2+N\lambda_WI)^{-1}-D\|^2_F+\frac{\lambda_H}2\|\Sigma\|^2_F+\frac{\lambda_W}2\|D\Sigma(\Sigma^2+N\lambda_WI)^{-1}\|^2_F\\
&\Leftrightarrow&\mathop{\arg\min}_{\sigma_i,i=1,\cdots,K}\frac1{2N}\sum^K_{i=1}n_i\left(\frac{\sigma^2_i}{\sigma^2_i+N\lambda_W}-1\right)^2+\frac{\lambda_H}2\sigma^2_i+\frac{\lambda_Wn_i}2\frac{\sigma^2_i}{(\sigma^2_i+N\lambda_W)^2}\\
&\Leftrightarrow&\mathop{\arg\min}_{\sigma_i,i=1,\cdots,K}\sum^K_{i=1}\frac{n_i\lambda_W}{2(\sigma^2_i+N\lambda_W)}+\frac{\lambda_H}2(\sigma^2_i+N\lambda_W)
\end{eqnarray*}
It is obvious that the above objective function is separable. By the property of the function $\frac{n_i\lambda_W}x+\lambda_Hx$, 
\[\sigma^*_i=\mathop{\arg\min}_{\sigma_i}\frac{n_i\lambda_W}{2(\sigma^2_i+N\lambda_W)}+\frac{\lambda_H}2(\sigma^2_i+N\lambda_W)=\left\{
\begin{array}{ll}
\sqrt{\sqrt{\frac{n_i\lambda_W}{\lambda_H}}-N\lambda_W},&\lambda_H\lambda_W\le\frac{n_i}{N^2},\\
0,&\hbox{otherwise}.
\end{array}
\right.\]
Define $\Sigma_*=\mathrm{diag}([\sigma^*_1\ \cdots \sigma^*_K])$, then 
\begin{equation*}
\begin{split}
W^*\bar H^*=&D\Sigma^2_*(\Sigma^2_*+N\lambda_WI)^{-1} D^{-1}=\left(D-N\sqrt{\lambda_W\lambda_H}I\right)_+D^{-1},\\
W^*(W^*)^\top=&D^2\Sigma^2_*(\Sigma^2_*+N\lambda_WI)^{-2}=\sqrt{\frac{\lambda_H}{\lambda_W}}\left(D-N\sqrt{\lambda_W\lambda_H}I\right)_+,\\ 
(\bar H^*)^\top\bar H^*=&\Sigma^2_*D^{-2}=\sqrt{\frac{\lambda_W}{\lambda_H}}\left(D-N\sqrt{\lambda_W\lambda_H}I\right)_+D^{-2}.
\end{split}
\end{equation*}
Note that $H^*=\bar H^*Y$, then
 \begin{equation*}
\begin{split}
W^* H^*=&W^*\bar H^*Y,\ 
( H^*)^\top H^*=Y^\top (\bar H^*)^\top\bar H^*Y.
\end{split}
\end{equation*}
\eproof

\subsection{Proof of Theorem \ref{theo:ufm-bias}} \label{proof:theo-ufm-bias}
\Proof
We consider
\[f(W,H,b)=\frac1{2N}\|WH+b\mathbbm{1}^\top_N-Y\|^2_F+\frac{\lambda_W}2\|W\|^2_F+\frac{\lambda_H}2\|H\|^2.\]
Take the partial derivative of $f$ about $b$ and set it to 0, we have 
\[\frac{\partial f}{\partial b}=\frac1N(WH+b\mathbbm{1}^\top_N-Y)\mathbbm{1}_N=0. \]%
That is,
\begin{equation}
\label{eq:b}
\begin{split}
b=&\frac1{N}(Y\mathbbm{1}_N-WH\mathbbm{1}_N)
=\frac1{N}\left([n_1,\cdots,n_K]^\top-W\sum^K_{k=1}\sum^{n_k}_{i=1}h_{k,i}\right)=\frac{\vn}N-Wh_G,%
\end{split}
\end{equation}
where $h_G=\frac1{N}\sum^K_{k=1}\sum^{n_k}_{i=1}h_{k,i}$ is the global centroid.
 Substitute $b$ in \eqref{eq:b} to $f(W,H,b)$ and get
\begin{equation}
\begin{split}
&f\left(W,H,\frac{\vn}N-Wh_G\right)\\=&\frac1{2N}\left\|W(H-h_G\mathbbm{1}^\top_N)-\left(Y-\frac{\vn}{N}\mathbbm{1}^\top_N\right)\right\|^2_F+\frac{\lambda_W}2\|W\|^2_F+\frac{\lambda_H}2\|H\|^2\\
=&\frac1{2N}\sum^K_{k=1}\sum^{n_k}_{i=1}\left\|W(h_{k,i}-h_G)-e_k+\frac{\vn}N\right\|^2_2+\frac{\lambda_W}2\|W\|^2_F+\frac{\lambda_H}2\sum^K_{k=1}\sum^{n_k}_{i=1}\|h_{k,i}\|^2.
\end{split}
\end{equation}
We take the derivative of $f\left(W,H,\frac{\vn}N-Wh_G\right)$ with respect to $h_{k,i}$ and set it to zero
\begin{equation}
\begin{split}
0=&\frac{\partial f}{\partial h_{k,i}}
=\frac1N\sum^K_{\ell=1}\sum^{n_\ell}_{j=1}\left(\delta_{k\ell,ij}-\frac{1}{N}\right)W^\top\left(W\left(h_{\ell,j}-h_G\right)-\left(e_k-\frac{\vn}{N}\right)\right)+\lambda_Hh_{k,i}\\
=&\frac 1N W^\top W(h_{k,i}-h_G)+\lambda_Hh_{k,i}-\frac{1}{N}W^\top e_k+\frac{1}{N}W^\top\frac{\vn}N,
\end{split}
\end{equation}
where $\delta_{k\ell,ij}=1$ if $k=\ell$ and $i=j$, 0 otherwise. 
Then
\[0=\frac1N\sum^K_{k=1}\sum^{n_k}_{i=1}\frac{\partial f}{\partial h_{k,i}}=\lambda_Hh_G,\]
which implies $h_G=0$. Hence, $b=\frac{\vn}{N}$ when $f(W,H,b)$ achieves critical points. Let $\tilde y_k=e_k-\frac{\vn}{N}$, define
\begin{equation}
\label{eq:tilde_f}
\begin{split}
&\tilde f(W,H)=f\left(W,H,\frac{\vn}{N}\right)\\
=&\frac1{2N}\sum^K_{k=1}n_k\frac1{n_k}\sum^{n_k}_{i=1}\|Wh_{k,i}-\tilde y_k\|^2_2+\frac{\lambda_W}2\|W\|^2_F+\frac{\lambda_H}2\sum^K_{k=1}n_k\frac1{n_k}\sum^{n_k}_{i=1}\|h_{k,i}\|^2\\
\ge&\frac1{2N}\sum^K_{k=1}n_k\left\|W\left(\frac1{n_k}\sum^{n_k}_{i=1}h_{k,i}\right)-\tilde y_k\right\|^2_2+\frac{\lambda_W}2\|W\|^2_F+\frac{\lambda_H}2\sum^K_{k=1}n_k\left\|\frac1{n_k}\sum^{n_k}_{i=1}h_{k,i}\right\|^2\\
=&\frac1{2N}\sum^K_{k=1}n_k\left\|W\bar h_k-\tilde y_k\right\|^2_2+\frac{\lambda_W}2\|W\|^2_F+\frac{\lambda_H}2\sum^K_{k=1}n_k\left\|\bar h_k\right\|^2_2\\
=&\frac1{2N}\|(W\bar H-\tilde Y)D\|^2_F+\frac{\lambda_W}2\|W\|^2_F+\frac{\lambda_H}2\|\bar HD\|^2_F.
\end{split}
\end{equation}
where $\bar h_k=\frac1{n_k}\sum^{n_k}_{i=1}h_{k,i}$, $\bar H=\begin{bmatrix}
    \bar h_1&\cdots&\bar h_K
\end{bmatrix}$ and $\tilde Y=\begin{bmatrix}
    \tilde y_1&\cdots&\tilde y_K
\end{bmatrix}$. Define $\hat f(W,\bar H)=\frac1{2N}\|(W\bar H-\tilde Y)D\|^2_F+\frac{\lambda_W}2\|W\|^2_F+\frac{\lambda_H}2\|\bar HD\|^2_F$. From \eqref{eq:tilde_f}, we have $\tilde f(W,H)\ge f_1(W,\bar H)$, and $\tilde f(W,H)=\hat f(W,\bar H)$ if and only if $h_{k,1}=\cdots=h_{k,n_k},\ k=1,\cdots,K$. Therefore,
\begin{equation}
\label{eq:model_z1}
\begin{split}
&\min_{W,\bar H}\ \hat f(W,\bar H)=\min_{W,\bar H}\ \frac1{2N}\|W\bar HD-\tilde Y D\|^2_F+\frac{\lambda_W}{2}\|W\|^2_F+\frac{\lambda_H}2\|\bar HD\|^2_F\\
=&\min_Z\ \left(\frac1{2N}\|Z-\tilde Y D\|^2_F+\min_{W,\bar H,Z=W\bar HD}\frac{\lambda_W}{2}\|W\|^2_F+\frac{\lambda_H}2\|\bar HD\|^2_F\right)\\
=&\min_Z\ \frac1{2N}\|Z-\tilde YD\|^2_F+\sqrt{\lambda_W\lambda_H}\|Z\|_*\triangleq \min_Z\ g(Z).
\end{split}
\end{equation}
Note that $\tilde YD=\tilde U \kappa\tilde V^\top$, then the minimizer of \eqref{eq:model_z1} is $Z^*=\tilde U( \kappa-N\sqrt{\lambda_W\lambda_H}I)_+\tilde V^\top$. To get the global minimizer of $\tilde f(W,H)$, we analyze the minimizer of $\hat f(W,\bar H)$ first. Then we obtain the optimal value of $H$ using the fact of $h_{k,1}=\cdots=h_{k,n_k},\ k=1,\cdots,K$ if and only if $\tilde f(W,H)=\hat f(W,\bar H)$. 

By simple calculation, we have
\begin{equation}
\begin{split}
\frac{\partial \hat f}{\partial W}=&\frac1N(W\bar H-\tilde Y)D^2\bar H^\top+\lambda_WW,\\
\frac{\partial \hat f}{\partial \bar H}=&\frac1NW^\top(W\bar H-\tilde Y)D^2+\lambda_H\bar HD^2.
\end{split}
\end{equation}
Set $\frac{\partial \hat f}{\partial W}=0$ and get $W(\bar HD^2\bar H^\top+N\lambda_WI)-\tilde YD^2\bar H^\top=0$. That is, $W=\tilde YD^2\bar H^\top(\bar HD^2\bar H^\top+N\lambda_WI)^{-1}$. Let $\bar HD=U\Sigma V^\top$ be the singular value decomposition, then $W=\tilde YD^2\bar H^\top(\bar HD^2\bar H^\top+N\lambda_WI)^{-1}=\tilde YDV\Sigma(\Sigma^2+N\lambda_WI)^{-1}U^\top$. 
Since
\begin{align*}
\tilde U(\kappa-N\sqrt{\lambda_W\lambda_H}I)_+\tilde V^\top=&Z^*=W\bar HD=\tilde YDV\Sigma^2(\Sigma^2+N\lambda_WI)^{-1}V^\top\\=&\tilde U\kappa\tilde V^\top V\Sigma^2(\Sigma^2+N\lambda_WI)^{-1}V^\top,
\end{align*}
which implies $V=\tilde V$. Note that $\tilde YD=\tilde U\kappa\tilde V^\top$. Then 
\begin{eqnarray*}
\hat f(W,\bar H)
&=&\frac{N\lambda_W^2}{2}\|\tilde YDV(\Sigma^2+N\lambda_WI)^{-1}\|^2_F+\frac{\lambda_H}2\|\Sigma\|^2_F+\frac{\lambda_W}2\|\tilde YDV\Sigma(\Sigma^2+N\lambda_WI)^{-1}\|^2_F\\
&=&\frac{1}{2}\sum^K_{k=1}\frac{\kappa^2_kN\lambda_W^2}{(\sigma^2_k+N\lambda_W)^2}+\frac{\kappa^2_k\sigma^2_k\lambda_W}{(\sigma^2_k+N\lambda_W)^2}+\frac{\lambda_H}2\sum^K_{k=1}\sigma^2_k\\
&=&\sum^K_{k=1}\frac{\kappa^2_k\lambda_W}{2(\sigma^2_k+N\lambda_W)}+\frac{\lambda_H}2(\sigma^2_k+N\lambda_W).
\end{eqnarray*}
Thus,
\begin{eqnarray*}
\mathop{\arg\min}_{W,\bar H}\hat f(W,\bar H)
\Leftrightarrow&\mathop{\arg\min}_{\sigma_k,k=1,\cdots,K}\sum^K_{k=1}\frac{\kappa^2_k\lambda_W}{2(\sigma^2_k+N\lambda_W)}+\frac{\lambda_H}2(\sigma^2_k+N\lambda_W).
\end{eqnarray*}
It is obvious that the above objective function is separable. By the property of the function $\frac{\sigma^2_k\lambda_W}x+\lambda_Hx$, 
\[\sigma^*_k=\mathop{\arg\min}_{\sigma_k}\frac{\kappa^2_k\lambda_W}{2(\sigma^2_k+N\lambda_W)}+\frac{\lambda_H}2(\sigma^2_k+N\lambda_W)=\left\{
\begin{array}{ll}
\sqrt{\sqrt{\frac{\kappa^2_k\lambda_W}{\lambda_H}}-N\lambda_W},&\lambda_H\lambda_W\le\frac{\kappa^2_k}{N^2},\\
0,&\hbox{otherwise}.
\end{array}
\right.\]
Define $\Sigma_*=\mathrm{diag}([\sigma^*_1\ \cdots\ \sigma^*_K])$, then 
\begin{equation*}
\begin{split}
W^*\bar H^*=&\tilde YD\tilde V\Sigma^2_*(\Sigma^2_*+N\lambda_WI)^{-1}\tilde V^\top=\tilde U\left(\kappa-N\sqrt{\lambda_W\lambda_H}\right)_+ \tilde V^\top D^{-1},\\
W^*(W^*)^\top=&\tilde YD\tilde V\Sigma^2_*(\Sigma^2_*+N\lambda_WI)^{-2}\tilde V^\top D\tilde Y^\top=\sqrt{\frac{\lambda_H}{\lambda_W}}\tilde U\left(\kappa-N\sqrt{\lambda_W\lambda_H}I\right)_+\tilde U^\top,\\ 
(\bar H^*)^\top\bar H^*=&D^{-1}\tilde V\Sigma^2_*\tilde VD^{-1}=\sqrt{\frac{\lambda_W}{\lambda_H}}D^{-1}\tilde V\left(\kappa-N\sqrt{\lambda_W\lambda_H}I\right)_+\tilde V^\top D^{-1}.
\end{split}
\end{equation*}
\eproof

\subsection{Proof of Theorem \ref{theo:bias_mult-layer}}\label{proof:bias-multilayer}
\Proof
We take the derivative of $f(W_L,\cdots,W_1,H,b)$ about $b$ and set $H_G=\frac1NH\mathbbm{1}_N$, then
\[\frac1N(W_L\cdots W_1H+b\mathbbm{1}^{\top}_N-Y)\mathbbm{1}_N=0\Leftrightarrow \frac1N(W_L\cdots W_1NH_G+Nb-\vn)=0\Leftrightarrow b=\frac{\vn}N-W_L\cdots W_1H_G.\]
Next we prove $H_G=0$.  Substitute $b$ into $f(W_L,\cdots,W_1,H,b)$ and get
\begin{equation}
\begin{split}
&f\left(W_L,\cdots,W_1,H,\frac{\vn}N-W_L\cdots W_1h_G\right)\\=&\frac1{2N}\left\|W_L\cdots W_1(H-h_G\mathbbm{1}^{\top}_N)-\left(Y-\frac{\vn}{N}\mathbbm{1}^{\top}_N\right)\right\|^2_F+\frac{1}{2}\sum^L_{j=1}\lambda_{W_j}\|W_j\|^2_F+\frac{\lambda_H}2\|H\|^2\\
=&\frac1{2N}\sum^K_{k=1}\sum^{n_k}_{i=1}\left\|W_L\cdots W_1(h_{k,i}-h_G)-e_k+\frac{\vn}N\right\|^2_2+\frac{1}{2}\sum^L_{j=1}\lambda_{W_j}\|W_j\|^2_F+\frac{\lambda_H}2\sum^K_{k=1}\sum^{n_k}_{i=1}\|h_{k,i}\|^2\\
\triangleq&\tilde f(W_L,\cdots,W_1,H,H_G).
\end{split}
\end{equation}
We take the derivative of $\tilde f(W_L,\cdots,W_1,H,H_G)$ with respect to $h_{k,i}$ and set it to zero
\begin{equation}
\begin{split}
0=&\frac{\partial \tilde f}{\partial h_{k,i}}\\
=&\frac1N\sum^K_{\ell=1}\sum^{n_\ell}_{j=1}\left(\delta_{k\ell,ij}-\frac{1}{N}\right)W^\top_1\cdots W^\top_L\left(W_L\cdots W_1\left(h_{\ell,j}-h_G\right)-\left(e_\ell-\frac{\vn}{N}\right)\right)+\lambda_Hh_{k,i}\\
=&\frac 1N W^\top_1\cdots W^\top_L W_L\cdots W_1(h_{k,i}-h_G)+\lambda_Hh_{k,i}-\frac{1}{N}W^\top_1\cdots W^\top_L e_k+\frac{1}{N}W^\top_1\cdots W^\top_L\frac{\vn}N,
\end{split}
\end{equation}
where $\delta_{k\ell,ij}=1$ if $k=\ell$ and $i=j$, 0 otherwise. 
Then
\[0=\frac1N\sum^K_{k=1}\sum^{n_k}_{i=1}\frac{\partial \tilde f}{\partial h_{k,i}}=\lambda_Hh_G.\]
Since $\lambda_H>0$, then  we can get $h_G=0$. Hence, $b=\frac{\vn}{N}$ when $f(W_L,\cdots,W_1,H,b)$ achieves critical points. Set $\tilde Y=I_K-\frac{\vn}{N}\mathbbm{1}^\top$. Then 
\begin{equation*}
\begin{split}
&\tilde f(W_L,\cdots,W_1,H,0)\\=&\frac1{2N}\|W_L\cdots W_1H-\tilde Y\|^2_F+\frac{1}{2}\sum^L_{j=1}\lambda_{W_j}\|W_j\|^2_F+\frac{\lambda_{H}}2\|H\|^2_F\\
=&\frac1{2N}\sum^K_{k=1}n_k\frac1{n_k}\sum^{n_k}_{i=1}\left\|W_L\cdots W_1h_i-e_k+\frac{\vn}N\right\|^2_2+\frac{1}{2}\sum^L_{j=1}\lambda_{W_j}\|W_j\|^2_F+\frac{\lambda_{H}}2\sum^K_{n=1}n_k\frac1{n_k}\sum^{n_k}_{i=1}\|h_i\|^2_2\\
\ge&\frac1{2N}\sum^K_{k=1}n_k\left\|W_L\cdots W_1\left(\frac1{n_k}\sum^{n_k}_{i=1}h_i\right)+\frac{\vn}N-e_k\right\|^2_2+\frac{1}{2}\sum^L_{j=1}\lambda_{W_j}\|W_j\|^2_F+\frac{\lambda_{H}}2\sum^K_{n=1}n_k\left\|\frac1{n_k}\sum^{n_k}_{i=1}h_i\right\|^2_2.
\end{split}
\end{equation*}
Set $\bar h_k=\frac1{n_k}\sum^{n_k}_{i=1}h_i$, $\bar H=[\bar h_1\ \cdots\ \bar h_K]$, then
\begin{equation*}
\begin{split}
f(W_L,\cdots,W_1,H,0)\ge&\frac1{2N}\sum^K_{k=1}n_k\left\|W_L\cdots W_1\bar h_k+\frac{\vn}{N}-e_k\right\|^2_2+\frac{1}{2}\sum^L_{j=1}\lambda_{W_j}\|W_j\|^2_F+\frac{\lambda_{H}}2\sum^K_{n=1}n_k\left\|\bar h_k\right\|^2_2\\
=&\frac1{2N}\|W_L\cdots W_1\bar HD-\tilde YD\|^2_F+\frac{1}{2}\sum^L_{j=1}\lambda_{W_j}\|W_j\|^2_F+\frac{\lambda_{H}}2\|\bar HD\|^2_F\\
=&\frac1{2N}\|W_L\cdots W_1\bar E-\hat Y\|^2_F+\frac{1}{2}\sum^L_{j=1}\lambda_{W_j}\|W_j\|^2_F+\frac{\lambda_{H}}2\|\bar E\|^2_F
\triangleq\hat f(W_L,\cdots,W_1,\bar E),
\end{split}
\end{equation*}
where $\bar E=\bar HD$ and $\hat Y=\tilde YD$. 
$\tilde f(W_L,\cdots,W_1,H,0)=\hat f(W_L,\cdots,W_1,\bar E)$ if and only if $h_{k,1}=\cdots=h_{k,n_k},\ k=1,\cdots,K$. Next, we describe the critical points of $\hat f(W_L,\cdots,W_1,\bar E)$. Any critical point of $\hat f(W_L,\cdots,W_1,\bar E)$ satisfies 
\begin{align*}
    \frac{\partial \hat f}{\partial W_L}=&\frac{1}{N}
(W_L\cdots W_1\bar E-\hat Y)\bar E^\top W_1^\top\cdots W_{L-1}^\top+\lambda_{W_L}W_L=0,\\
\cdots&\\
    \frac{\partial \hat f}{\partial W_{\ell}}=&\frac{1}{N}W_{\ell+1}^\top\cdots W_L^\top(W_L\cdots W_1\bar E-\hat Y)\bar E^\top W^\top_1\cdots W_{\ell-1}^\top+\lambda_{W_\ell}W_\ell=0,\\
    \frac{\partial \hat f}{\partial W_{\ell-1}}=&\frac{1}{N}W_{\ell}^\top\cdots W_L^\top(W_L\cdots W_1\bar E-\hat Y)\bar E^\top W^\top_1\cdots W_{\ell-2}^\top+\lambda_{W_{\ell-1}}W_{\ell-1}=0,\\
    \cdots&\\
    \frac{\partial \hat f}{\partial W_1}=&\frac{1}{N}W_2^\top\cdots W_{L-1}^\top
(W_L\cdots W_1\bar E-\hat Y)\bar E^\top +\lambda_{W_1}W_1=0,\\
    \frac{\partial \hat f}{\partial \bar E}=&\frac{1}{N}W_1^\top\cdots W_{L-1}^\top
    (W_L\cdots W_1\bar E-\hat Y) +\lambda_{H}\bar E=0
\end{align*}
Hence, we have $W^\top_\ell\frac{\partial f}{\partial W_{\ell}}=\frac{\partial f}{\partial W_{\ell-1}}W^\top_{\ell-1}=0,\forall \ell=2,\cdots,L$, which implies $\lambda_{W_\ell}W^\top_\ell W_{\ell}=\lambda_{W_{\ell-1}}W_{\ell-1} W^\top_{\ell-1}$ for $\ell=2,\cdots,L$. Therefore, we can conclude that for any global minimizer $(W_L,\cdots,W_1,\bar E)$ of $\hat f$, $\rank(W_1)=\cdots=\rank(W_L)$. Let $W_1=U_{W_1}\Sigma_{W_1}V^\top_{W_1}$, then
\[\lambda_{W_2}W^\top_2 W_{2}=\lambda_{W_{1}}W_{1} W^\top_{1}=\lambda_{W_1}U_{W_1}\Sigma^2_{W_1}U_{W_1}^\top\Rightarrow W_2=\sqrt{\frac{\lambda_{W_1}}{\lambda_{W_2}}}U_{W_2}\Sigma_{W_1}U_{W_1}^\top, \rank(W_2)=\rank(W_1),\]
where $U_{W_2}$ is an orthonormal matrix. Similarly, we have the following SVD decomposition 
\begin{align*}
    W_\ell=U_{W_\ell}\Sigma_{W_\ell}U^\top_{W_{\ell-1}},\ell=2,\cdots,L,\ \bar E=V_{W_1}\Sigma_{\bar E}V^\top_{\bar E},
\end{align*}
where 
\[\Sigma_{W_j}=\sqrt{\frac{\lambda_{W_1}}{\lambda_{W_j}}}\Sigma_{W_1},j=1,\cdots,L,\ \Sigma_{\bar E}=\sqrt{\frac{\lambda_{W_1}}{\lambda_{H}}}\Sigma_{W_1},\]
and $U_{W_j},j=1,\cdots,L$, $V_{W_1}$, $V_{\bar E}$ are all orthonormal matrices. Then we can claim that \[U_{W_L}=\hat U,\ V_{\bar E} = \hat V,\]
where $\hat U, \hat V$ are the left and right orthonormal matrices of $\hat Y$. That is $\hat Y=\hat U \kappa\hat V^\top$. Otherwise, let $(W'_L,\cdots,W'_1,\bar E')$ be a global minimizer of $f'$ and the SVD of $W'_L=U_{W'_L}\Sigma_{W'_L}U_{W'_{L-1}}^\top$, $\bar E'=V_{W'_1}\Sigma_{\bar E'}V_{\bar E'}^\top$, but $U_{W'_L}\neq\hat U$ or $V_{\bar E'} \neq \hat V$. Let $W''_L=\hat U\Sigma_{W'_L}U_{W'_{L-1}}^\top$, $\bar E''=V_{W'_1}\Sigma_{\bar E'}\hat V^\top$ then $\hat f(W'_L,\cdots,W'_1,\bar E')>\hat f(W''_L,W'_{L-1},\cdots,W'_1,\bar E'')$, which conflicts with the definition of a global minimizer. 

Set $c=\frac{\lambda^{L-1}_{W_1}}{\lambda_{W_L}\lambda_{W_{L-1}}\cdots\lambda_{W_2}}$, $r=\min(d_L,\cdots,d_1,K)$ and $\Sigma_{W_1}=\begin{bmatrix}
    \diag(\sigma_1,\cdots,\sigma_r)&0\\
    0&0
\end{bmatrix}$.  Substitute the forms of SVD of $W_1,\cdots, W_L$ into 
\[\frac{\partial \hat f}{\partial \bar E}=\frac{1}{N}W_1^\top\cdots W_{L-1}^\top W_L^\top
(W_L\cdots W_1\bar E-\hat Y) +\lambda_H\bar E=0,\]
then 
\begin{align*}
    \bar E=&U_{W_1}\begin{bmatrix}
    \diag\left(\frac{\sqrt{c}\sigma^L_1}{c\sigma_1^{2L}+N\lambda_H},\cdots,\frac{\sqrt{c}\sigma^L_r}{c\sigma_r1^{2L}+N\lambda_H}\right)&0_{r\times (K-r)}\\
    0_{(d_1-r)\times r}&0_{(d_1-r)\times (K-r)}\\
\end{bmatrix}U_{W_L}^\top\hat Y\\
=&U_{W_1}\begin{bmatrix}
    \diag\left(\frac{\sqrt{c}\sigma^L_1}{c\sigma_1^{2L}+N\lambda_H},\cdots,\frac{\sqrt{c}\sigma^L_r}{c\sigma_r1^{2L}+N\lambda_H}\right)&0_{r\times (K-r)}\\
    0_{(d_1-r)\times r}&0_{(d_1-r)\times (K-r)}\\
\end{bmatrix}\kappa\hat V^\top,
\end{align*}
where $\kappa=\diag([\kappa_1\ \cdots\ \kappa_K])$ is the singular value matrix of $\hat Y$ and
\begin{equation*}
    W_L\cdots W_1\bar E-\hat Y=\hat U\begin{bmatrix}
    \diag\left(\frac{-N\lambda_H}{c\sigma_1^{2L}+N\lambda_H},\cdots,\frac{-N\lambda_H}{c\sigma_r1^{2L}+N\lambda_H}\right)&0_{r\times (K-r)}\\
    0_{(d_L-r)\times r}&-I_{(d_L-r)\times (K-r)}\\
\end{bmatrix}\kappa \hat V^\top.
\end{equation*}
Then we have 
\begin{equation}
\begin{split}
&\hat f(W_L,\cdots,W_1,\bar E)=\frac{1}{2N}\left\|W_L\cdots W_1\bar E-\hat Y\right\|^2_F+\sum^L_{i=1}\frac{\lambda_{W_i}}2\|W_i\|^2_F+\frac{\lambda_{H}}2\|\bar E\|^2_F\\
=&\frac{1}{2N}\sum^r_{j=1}\frac{\kappa^2_jN^2\lambda^2_H}{(c\sigma^{2L}_j+N\lambda_H)^2}+\frac{L\lambda_W}{2}\sum^r_{j=1}\sigma^2_j+\frac{\lambda_{H}}{2}\sum^r_{j=1}\frac{c\sigma^{2L}_j\kappa_j^2}{(c\sigma^{2L}_j+N\lambda_{H})^2}+\frac{1}{2N}\sum^K_{j=r+1}\kappa^2_j\\
=&\frac{1}{2}\sum^r_{j=1}\left(\frac{\lambda_H\kappa^2_j}{c\sigma^{2L}_j+N\lambda_{H}}+L\lambda_{W_1}\sigma^2_j\right).
\end{split}
\end{equation}
According to Lemma \ref{lem:min_g} by choosing $x=\sqrt[L]{\frac{c}{N\lambda_H}}\sigma^2_j$ and $\alpha=\frac{LN\lambda_{W_1}}{\kappa^2_j}\sqrt[L]{\frac{N\lambda_H}{c}}=\frac{LN}{\kappa_j^2}\sqrt[L]{N\lambda_{W_L}\cdots\lambda_{W_1}\lambda_H}$ for any $j$, we can conclude the result.
\eproof
\subsection{Proof of Theorem \ref{theo:singular}}
\label{subsec:proof_singular}
\begin{theo}\label{theo:singular_proof}
    We consider a dataset is with $(\{N_i,\ell_i\}^m_{i=1},\{n_j\}^K_{j=1})$-Imbalances  defined in Definition \ref{def:imbalance}. Without loss of generality we assume $\vn=\begin{bmatrix}
        n_1&\cdots&n_K
    \end{bmatrix}^\top=\begin{bmatrix}
    N_1\mathbbm{1}^\top_{\ell_1}&\cdots&N_m\mathbbm{1}^\top_{\ell_m}
    \end{bmatrix}^\top$ and $\sqrt{\vn}=\begin{bmatrix}
        \sqrt{N_1}\mathbbm{1}^\top_{\ell_1}&\cdots&\sqrt{N_m}\mathbbm{1}^\top_{\ell_m}
    \end{bmatrix}^\top$. Let $ D=\mathrm{diag}(\sqrt{\vn})$, $N=\sum^K_{j=1}n_j$ and $\tilde Y=(I_K-1/N\vn\mathbbm{1}^\top_K)D=D-\frac{\vn}{N}\sqrt{\vn}^\top$. Set $$G=\begin{bmatrix}
    \sqrt{N_1}(1-\frac{N_1\ell_1}{N})&-\frac{N_1\sqrt{N_2}}{N}\sqrt{\ell_1\ell_2}&\cdots&-\frac{N_1\sqrt{N_m}}{N}\sqrt{\ell_1\ell_m}\\
    -\frac{N_2\sqrt{N_1}}{N}\sqrt{\ell_2\ell_1}&\sqrt{N_2}(1-\frac{N_2\ell_2}{N})&\cdots&-\frac{N_2\sqrt{N_1}}{N}\sqrt{\ell_2\ell_m}\\
    \vdots&\vdots&\ddots&\vdots\\
    -\frac{N_m\sqrt{N_1}}{N}\sqrt{\ell_m\ell_1}&-\frac{N_m\sqrt{N_2}}{N}\sqrt{\ell_m\ell_2}&\cdots&\sqrt{N_m}(1-\frac{N_m\ell_m}{N})\\
\end{bmatrix},$$ and $\tilde\sigma_j,j=1,\cdots,m$ are the singular values of $G$ satisfying $\tilde\sigma_1\ge\cdots\ge\tilde\sigma_m$. Then the singular values of $\tilde Y$ satisfy 
\[s_i=\left\{
\begin{array}{ll}
    \sqrt{N_1}, & i=1,\cdots,\ell_1-1, \\
    \sqrt{N_j}, & i=\sum^j_{\alpha=1}\ell_{\alpha-1}+1,\cdots,\sum^j_{\alpha=1}\ell_\alpha-1,\ j=2,\cdots,m,\\
    \tilde\sigma_j, & i=\sum^j_{\alpha=1}\ell_\alpha,\ j=1,\cdots,m,
\end{array}
\right.\]
and $0=\tilde\sigma_m<\sqrt{N_m}<\tilde\sigma_{m-1}<\sqrt{N_{m-1}}<\cdots<\sqrt{N_2}<\tilde\sigma_1<\sqrt{N_1}$.

\end{theo}
\Proof
Denote $\tilde Y=(\tilde Y_{ij})^m_{i,j=1}$, where 
\begin{align*}
    \tilde Y_{ij} = \left\{
    \begin{array}{cc}
      \sqrt{N_i}(I_{\ell_i}-\frac{N_i}{N}\mathbbm{1}_{\ell_i}\mathbbm{1}^\top_{\ell_i}),   & i=j, \\
      -\frac{N_i\sqrt{N_j}}{N}\mathbbm{1}_{\ell_i}\mathbbm{1}^\top_{\ell_j},   & i\ne j. 
    \end{array}
    \right.
\end{align*}
Note that $\tilde Y_{ii}$ is symmetric, so its eigenvalues are real. It is evident that $\sqrt{N_i}\left(1-\frac{N_i\ell_i}{N}\right)$ is an eigenvalue of $\tilde Y_{ii}$. The corresponding eigenvector is $\frac{\mathbbm{1}_{\ell_i}}{\|\mathbbm{1}_{\ell_i}\|}$. 
Additionally, the remaining eigenvalues of $\tilde Y_{ii}$ are all $\sqrt{N_i}$, and their corresponding eigenvectors are $u^{(i)}_p,p=1,\cdots,\ell_i-1$, which are orthogonal to $\mathbbm{1}_{\ell_i}$. Then the eigenvalue decomposition of $\tilde Y_{ii}$ is $\tilde Y_{ii}=U_i\Sigma_{ii}U^\top_i$, where 
$\Sigma_{ii}=\mathrm{diag}([\sqrt{N_i}\ \cdots\ \sqrt{N_i}\ \sqrt{N_i}\left(1-\frac{N_i\ell_i}{N}\right)])$ and $U_i=\begin{bmatrix}
    u^{(i)}_1&\cdots&u^{(i)}_{\ell_i-1}&\frac{\mathbbm{1}_{\ell_i}}{\|\mathbbm{1}_{\ell_i}\|}
\end{bmatrix}$.

For $\tilde Y_{ij},i\ne j$, it is a rank-1 matrix, whose nonzero singular value is only $-\frac{N_i\sqrt{N_j}}{N}\sqrt{\ell_i\ell_j}$ and the corresponding left and right singular vectors are $\frac{\mathbbm{1}_{\ell_i}}{\|\mathbbm{1}_{\ell_i}\|}$ and $\frac{\mathbbm{1}_{\ell_j}}{\|\mathbbm{1}_{\ell_j}\|}$, respectively. Therefore, the singular value decomposition of $\tilde Y_{ij}$ can be represented as $\tilde Y_{ij}=U_i\Sigma_{ij}U^\top_j$ with $\Sigma_{ij}=\mathrm{diag}([0,\cdots,0,-\frac{N_i\sqrt{N_j}}{N}\sqrt{\ell_i\ell_j}])$. Define \[U=\begin{bmatrix}
    U_1&\cdots&0\\
    \vdots&\ddots&\vdots\\
    0&\cdots&U_m
\end{bmatrix},\ \Sigma=\begin{bmatrix}
    \Sigma_{11}&\cdots&\Sigma_{1m}\\
    \vdots&\ddots&\vdots\\
    \Sigma_{m1}&\cdots&\Sigma_{mm}
\end{bmatrix},\]
then
\begin{align*}
    \tilde Y=\begin{bmatrix}
    U_1\Sigma_{11}U^\top_1&U_1\Sigma_{12} U^\top_2&\cdots&U_1\Sigma_{1m}U^\top_m\\
    U_2\Sigma_{21}U^\top_1&U_2\Sigma _{22}U^\top_2&\cdots&U_2\Sigma_{2m}U^\top_m\\
    \vdots&\vdots&\ddots&\vdots\\
    U_m\Sigma_{m1}U^\top_1&U_m\Sigma _{m2}U^\top_2&\cdots&U_1\Sigma_{mm}U^\top_m\\
    \end{bmatrix}= U\Sigma U^\top.
\end{align*}

We take out the $\sum^{j}_{i=1}\ell_i$-th rows and columns of $\Sigma$ with $j=1,\cdots,m$ and form a new matrix $$G=\begin{bmatrix}
    \sqrt{N_1}(1-\frac{N_1\ell_1}{N})&-\frac{N_1\sqrt{N_2}}{N}\sqrt{\ell_1\ell_2}&\cdots&-\frac{N_1\sqrt{N_m}}{N}\sqrt{\ell_1\ell_m}\\
    -\frac{N_2\sqrt{N_1}}{N}\sqrt{\ell_2\ell_1}&\sqrt{N_2}(1-\frac{N_2\ell_2}{N})&\cdots&-\frac{N_2\sqrt{N_1}}{N}\sqrt{\ell_2\ell_m}\\
    \vdots&\vdots&\ddots&\vdots\\
    -\frac{N_m\sqrt{N_1}}{N}\sqrt{\ell_m\ell_1}&-\frac{N_m\sqrt{N_2}}{N}\sqrt{\ell_m\ell_2}&\cdots&\sqrt{N_m}(1-\frac{N_m\ell_m}{N})\\
\end{bmatrix}.$$ 
Let $G=\alpha^\top\tilde\Sigma\beta$, where 
\[\alpha=\begin{bmatrix}
    \alpha_{11}&\cdots&\alpha_{1m}\\
    \vdots&\ddots&\vdots\\
    \alpha_{m1}&\cdots&\alpha_{mm}
\end{bmatrix},\tilde\Sigma=\mathrm{diag}([\tilde\sigma_1\ \cdots\ \tilde\sigma_m]),\ \beta=\begin{bmatrix}
    \beta_{11}&\cdots&\beta_{1m}\\
    \vdots&\ddots&\vdots\\
    \beta_{m1}&\cdots&\beta_{mm}
\end{bmatrix}.\]
Define \[P=\begin{bmatrix}
    I_{\ell_1-1}&0&0&0&0&\cdots&0&0\\
    0&\alpha_{11}&0&\alpha_{12}&0&\cdots&0&\alpha_{1m}\\
    0&0&I_{\ell_2-1}&0&0&\cdots&0&0\\
    \vdots&\vdots&\vdots&\vdots&\vdots&\ddots&\vdots&\vdots\\
    0&0&0&0&0&\cdots&I_{\ell_m-1}&0\\
    0&\alpha_{m1}&0&\alpha_{m2}&0&\cdots&0&\alpha_{mm}\\
\end{bmatrix},\ \Lambda=\begin{bmatrix}
    \sqrt{N_1}I_{\ell_1-1}&0&\cdots&0&0\\
    0&\tilde\sigma_1&\cdots&0&0\\
    \vdots&\vdots&\ddots&\vdots&\vdots\\
    0&0&\cdots&\sqrt{N_m}I_{\ell_m-1}&0\\
    0&0&\cdots&0&\tilde\sigma_m
\end{bmatrix},\]
\[Q=\begin{bmatrix}
    I_{\ell_1-1}&0&0&0&0&\cdots&0&0\\
    0&\beta_{11}&0&\beta_{12}&0&\cdots&0&\beta_{1m}\\
    0&0&I_{\ell_2-1}&0&0&\cdots&0&0\\
    \vdots&\vdots&\vdots&\vdots&\vdots&\ddots&\vdots&\vdots\\
    0&0&0&0&0&\cdots&I_{\ell_m-1}&0\\
    0&\beta_{m1}&0&\beta_{m2}&0&\cdots&0&\beta_{mm}\\
\end{bmatrix},\]
then $P\Sigma Q^\top=\Lambda$. 
Hence, $\tilde Y=U\Sigma V^\top=UP^\top\Lambda QU^\top=(UP^\top)\Lambda(UQ^\top)^\top$.

Next, we will show $0=\tilde\sigma_m<\sqrt{N_m}<\tilde\sigma_{m-1}<\sqrt{N_{m-1}}<\cdots<\sqrt{N_2}<\tilde\sigma_1<\sqrt{N_1}$. Note that
\[GG^\top=\begin{bmatrix}
    \frac{N_1}{N}(N-N_1\ell_1)&-\frac{N_1N_2}{N}\sqrt{\ell_1\ell_2}&\cdots&-\frac{N_1N_m}{N}\sqrt{\ell_1\ell_m}\\
    -\frac{N_2N_1}{N}\sqrt{\ell_2\ell_1}&\frac{N_2}{N}(N-N_2\ell_2)&\cdots&-\frac{N_2N_m}{N}\sqrt{\ell_2\ell_m}\\
    \vdots&\vdots&\ddots&\vdots\\
    -\frac{N_mN_1}{N}\sqrt{\ell_m\ell_1}&-\frac{N_mN_2}{N}\sqrt{\ell_m\ell_2}&\cdots&
    \frac{N_m}{N}(N-N_m\ell_m)
\end{bmatrix}.\]
Define
\[f(\lambda)=|\lambda I-GG^\top|=\begin{vmatrix}
    \lambda-\frac{N_1}{N}(N-N_1\ell_1)&\frac{N_1N_2}{N}\sqrt{\ell_1\ell_2}&\cdots&\frac{N_1N_m}{N}\sqrt{\ell_1\ell_m}\\
    \frac{N_2N_1}{N}\sqrt{\ell_2\ell_1}&\lambda-\frac{N_2}{N}(N-N_2\ell_2)&\cdots&\frac{N_2N_m}{N}\sqrt{\ell_2\ell_m}\\
    \vdots&\vdots&\ddots&\vdots\\
    \frac{N_mN_1}{N}\sqrt{\ell_m\ell_1}&\frac{N_mN_2}{N}\sqrt{\ell_m\ell_2}&\cdots&
    \lambda-\frac{N_m}{N}(N-N_m\ell_m)
\end{vmatrix},\]
then the roots of $f(\lambda)=0$ are the eigenvalues of $GG^\top$. By simple calculation we have $f(0)=0$, $f(N_i)=\frac{N_i^2\ell_i}{N}\Pi_{j\ne i}(N_i-N_j)$, then the sign of $f(N_i)$ is $(-1)^{i-1},i=1,\cdots,m$. That is, $f(N_1)>0$, $f(N_2)<0$, $f(N_3)>0,\ \cdots,$ and so on. Therefore, there must exist zeros of $f$ in the interval $(N_i,N_{i+1}), i=1,\cdots,m-1$ and $0$ is the smallest root of $f(\lambda)=0$. Hence, $0=\tilde\sigma_m<\sqrt{N_m}<\tilde\sigma_{N-1}<\sqrt{n_{N-1}}<\cdots<\sqrt{N_2}<\tilde\sigma_1<\sqrt{N_1}$.
\eproof

\begin{theo}
    \label{theo:singular_proof1}
    We consider a dataset is with $(\{N_i,\ell_i\}^m_{i=1},\{n_j\}^K_{j=1})$-Imbalances  defined in Definition \ref{def:imbalance}. 
    Let $$G=\begin{bmatrix}
    \sqrt{N_1}(1-\frac{N_1\ell_1}{N})&-\frac{N_1\sqrt{N_2}}{N}\sqrt{\ell_1\ell_2}&\cdots&-\frac{N_1\sqrt{N_m}}{N}\sqrt{\ell_1\ell_m}\\
    -\frac{N_2\sqrt{N_1}}{N}\sqrt{\ell_2\ell_1}&\sqrt{N_2}(1-\frac{N_2\ell_2}{N})&\cdots&-\frac{N_2\sqrt{N_1}}{N}\sqrt{\ell_2\ell_m}\\
    \vdots&\vdots&\ddots&\vdots\\
    -\frac{N_m\sqrt{N_1}}{N}\sqrt{\ell_m\ell_1}&-\frac{N_m\sqrt{N_2}}{N}\sqrt{\ell_m\ell_2}&\cdots&\sqrt{N_m}(1-\frac{N_m\ell_m}{N})\\
\end{bmatrix}.$$
    Then
    \begin{itemize}
        \item when $m=2$, the singular values of $G$ are $0$ and $\sqrt{\frac{KN_1N_2}{N}}$.
        \item when $m=3$, the singular values of $G$ are $0$ and the roots of quadratic polynomial $\lambda^2-a\lambda+b=0$, where 
        $a = \frac 1N[N_1(N_2\ell_2+N_3\ell_3)+n_2(N_1\ell_1+N_3\ell_3)+N_3(N_1\ell_1+N_2\ell_2)]$, and $b = \frac{KN_1N_2N_3}{N}$.
    \end{itemize}
\end{theo}
\Proof
When $m=2$, we have
\begin{align*}
    GG^\top=\frac{N_1N_2}{N^2}\begin{bmatrix}
        \sqrt{N_2}\ell_2 & -\sqrt{N_1}\sqrt{\ell_1\ell_2}\\
        -\sqrt{N_2}\sqrt{\ell_1\ell_2}&\sqrt{N_1}\ell_1
    \end{bmatrix}
    \begin{bmatrix}
        \sqrt{N_2}\ell_2&-\sqrt{N_2}\sqrt{\ell_1\ell_2}\\
        -\sqrt{N_1}\sqrt{\ell_1\ell_2}&\sqrt{N_1}\ell_1
    \end{bmatrix}=\frac{N_1N_2}{N}\begin{bmatrix}
        \ell_2&-\sqrt{\ell_1\ell_2}\\
        -\sqrt{\ell_1\ell_2}&\ell_1
    \end{bmatrix}.
\end{align*}
Next we calculate the determinant of $GG^\top$ and the roots of $0=\hbox{det}(\lambda I-GG^\top)=\lambda^2-\frac{N_1N_2K}{N}\lambda$ are $0$ and $\frac{N_1N_2K}{N}$. Therefore, the singular values of $G$ are $0$ and $\sqrt{\frac{KN_1N_2}{N}}$.

When $m=3$, 
\begin{align*}
    GG^\top=\begin{bmatrix}
        \frac{N_1}{N}(N_2\ell_2+N_3\ell_3)&-\frac{N_1N_2}{N}\sqrt{\ell_1\ell_2}&-\frac{N_1N_3}{N}\sqrt{\ell_1\ell_3}\\
        -\frac{N_1N_2}{N}\sqrt{\ell_1\ell_2}&\frac{N_2}{N}(N_1\ell_1+N_3\ell_3)&-\frac{N_2N_3}{N}\sqrt{\ell_2\ell_3}\\
        -\frac{N_1N_3}{N}\sqrt{\ell_1\ell_3}&-\frac{N_2N_3}{N}\sqrt{\ell_2\ell_3}&\frac{N_3}{N}(N_1\ell_1+N_2\ell_2)\\
    \end{bmatrix}
\end{align*}
and 
\begin{align*}
    &\hbox{det}(\lambda I-GG^\top)\\
    =&\frac{1}{N^3}\begin{vmatrix}
        N\lambda-N_1(N_2\ell_2+N_3\ell_3)&N_1N_2\sqrt{\ell_1\ell_2}&N_1N_3\sqrt{\ell_1\ell_3}\\
        N_1N_2\sqrt{\ell_1\ell_2}&N\lambda-N_2(N_1\ell_1+N_3\ell_3)&N_2N_3\sqrt{\ell_2\ell_3}\\
        N_1N_3\sqrt{\ell_1\ell_3}&N_2N_3\sqrt{\ell_2\ell_3}& N\lambda-N_3(N_1\ell_1+N_2\ell_2)
    \end{vmatrix}\\
    =&\lambda^3-a\lambda^2+b\lambda=(\lambda^2-a\lambda+b)\lambda,
\end{align*}
we can easily get the roots of $\hbox{det}(\lambda I-GG^\top)=0$.
\eproof

\begin{figure}[ht]
	\begin{center}
		\subfigure{
			\includegraphics[width=4.5cm, height=4cm]{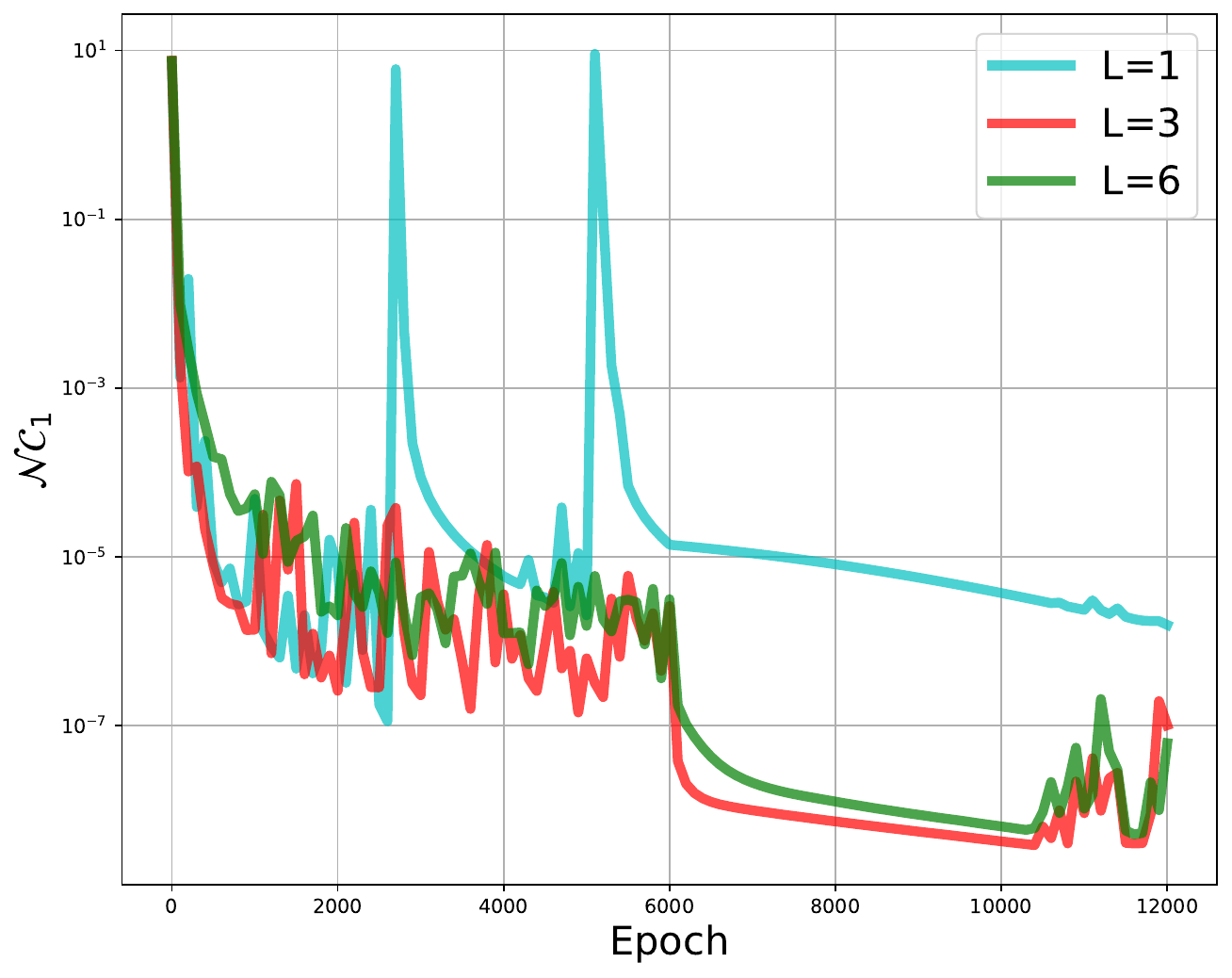}}
		\subfigure{
			\includegraphics[width=4.5cm, height=4cm]{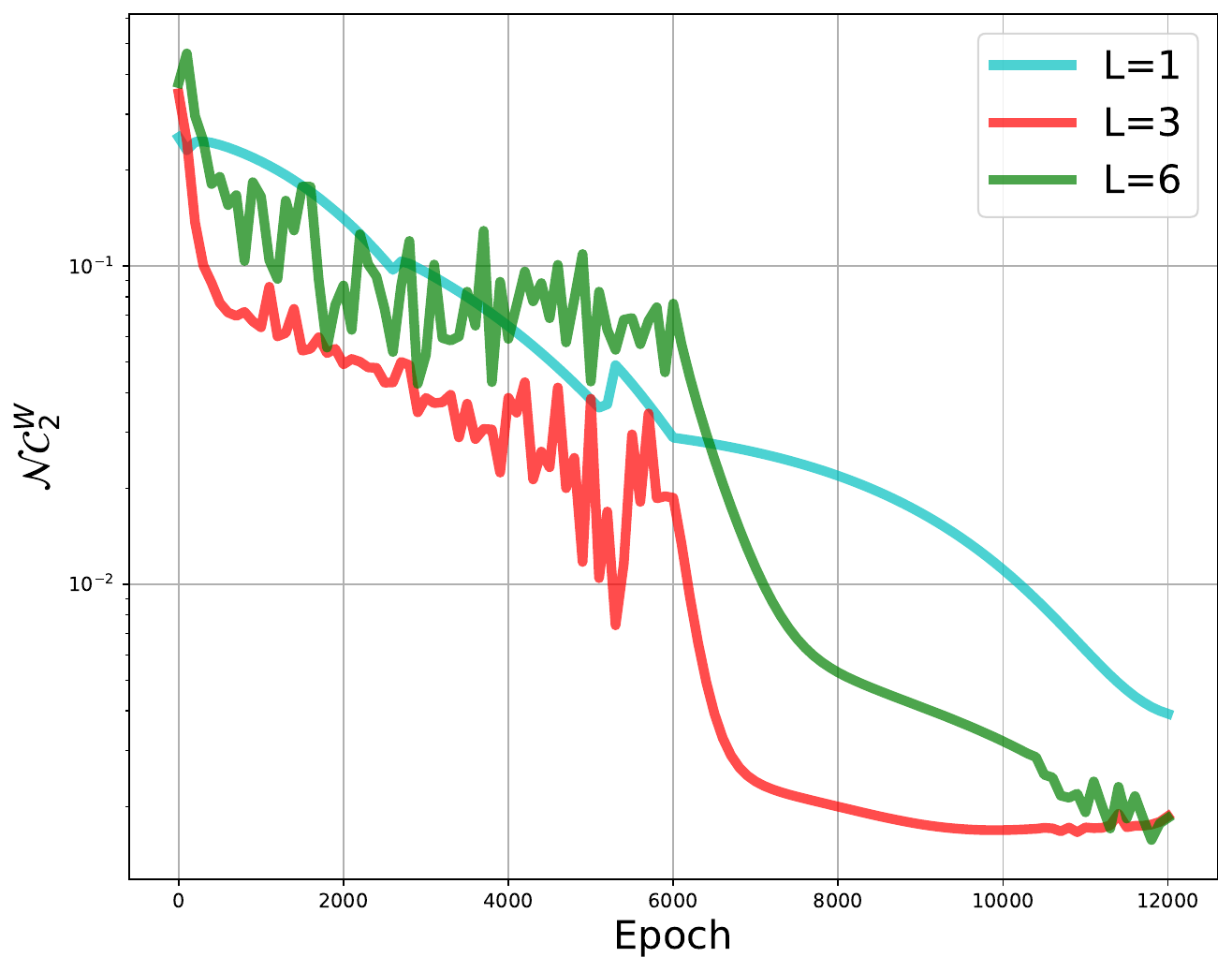}}
        \subfigure{
			\includegraphics[width=4.5cm, height=4cm]{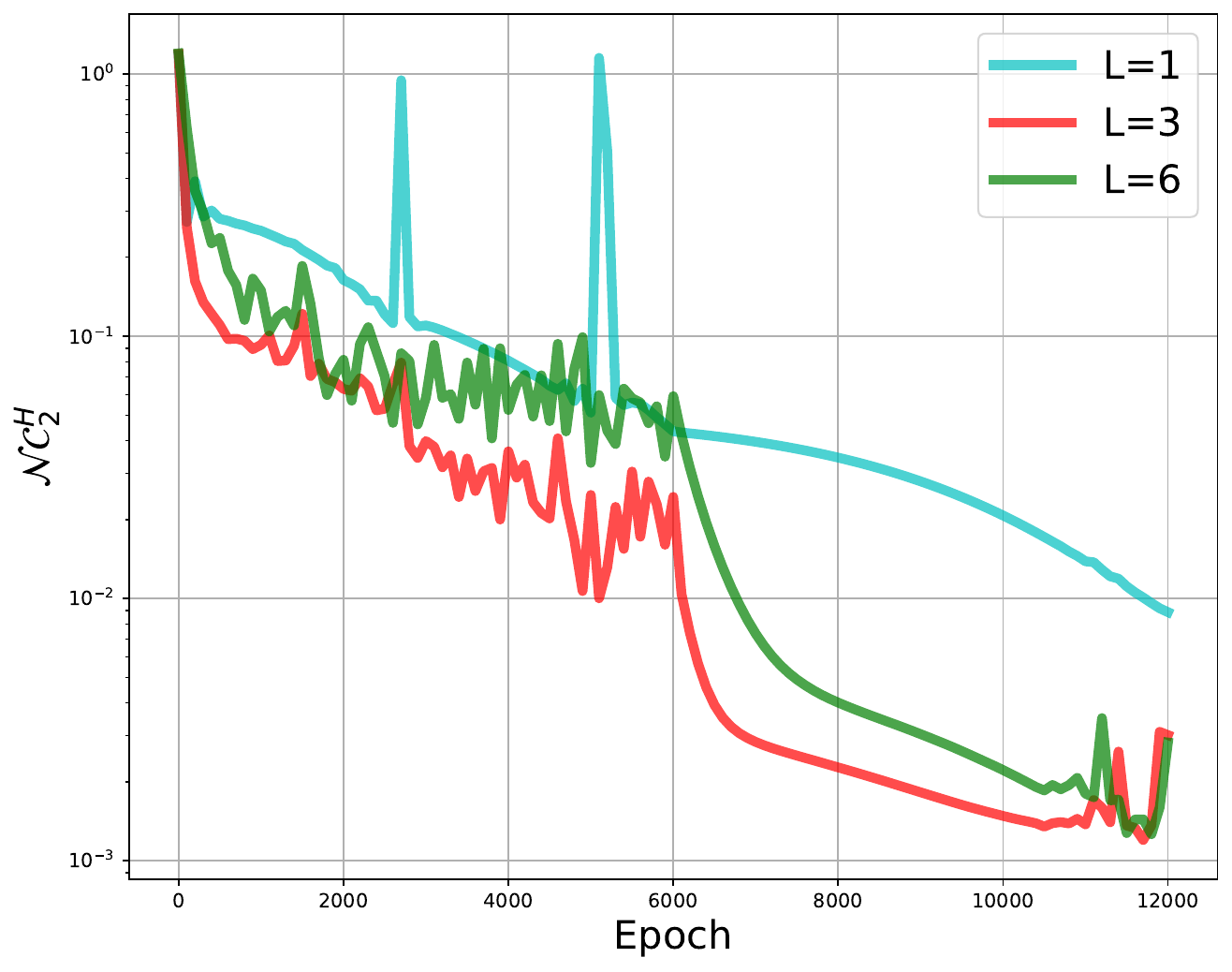}}
        \subfigure{
			\includegraphics[width=4.5cm, height=4cm]{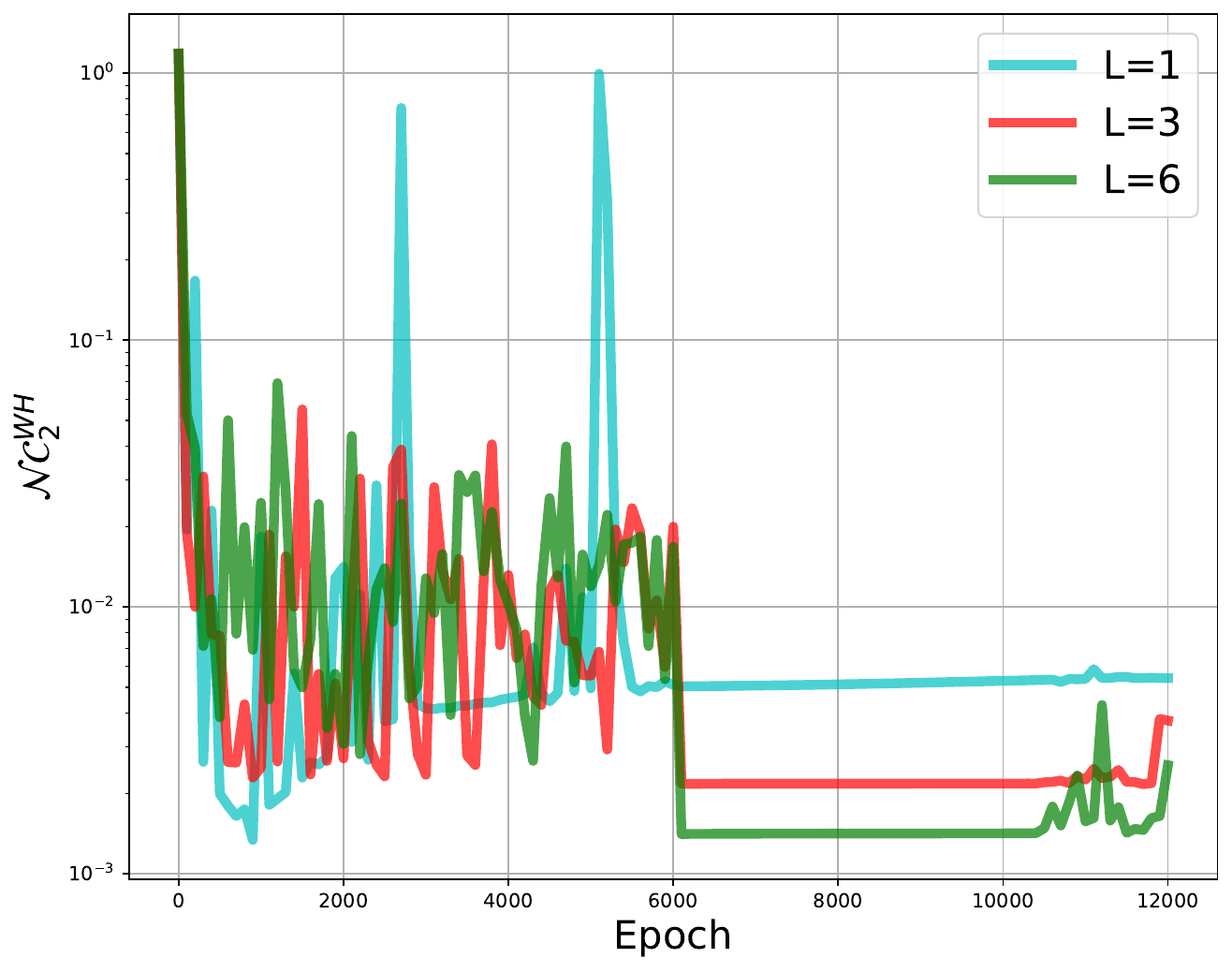}}
        \subfigure{
			\includegraphics[width=4.5cm, height=4cm]{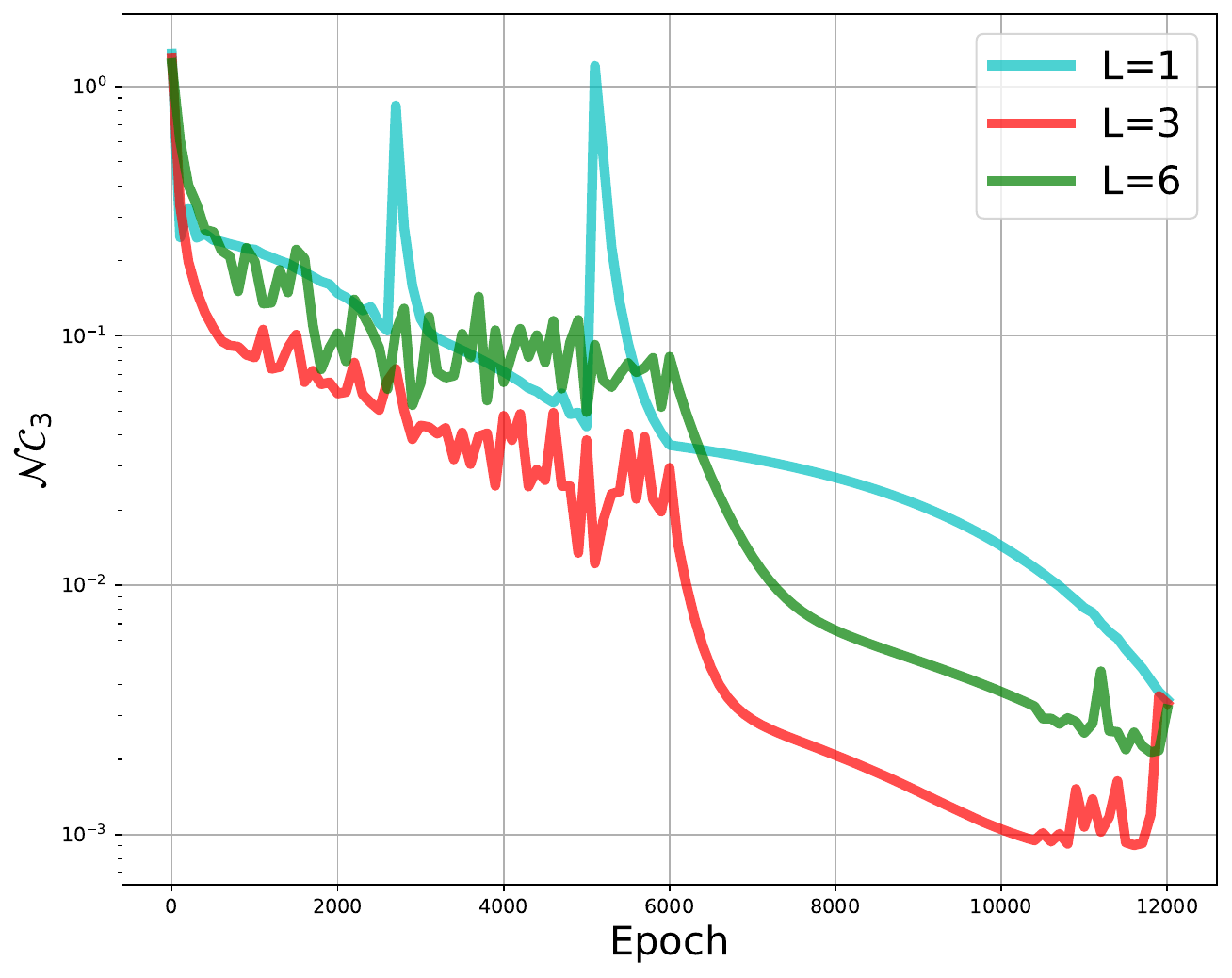}}
        \subfigure{
			\includegraphics[width=4.5cm, height=4cm]{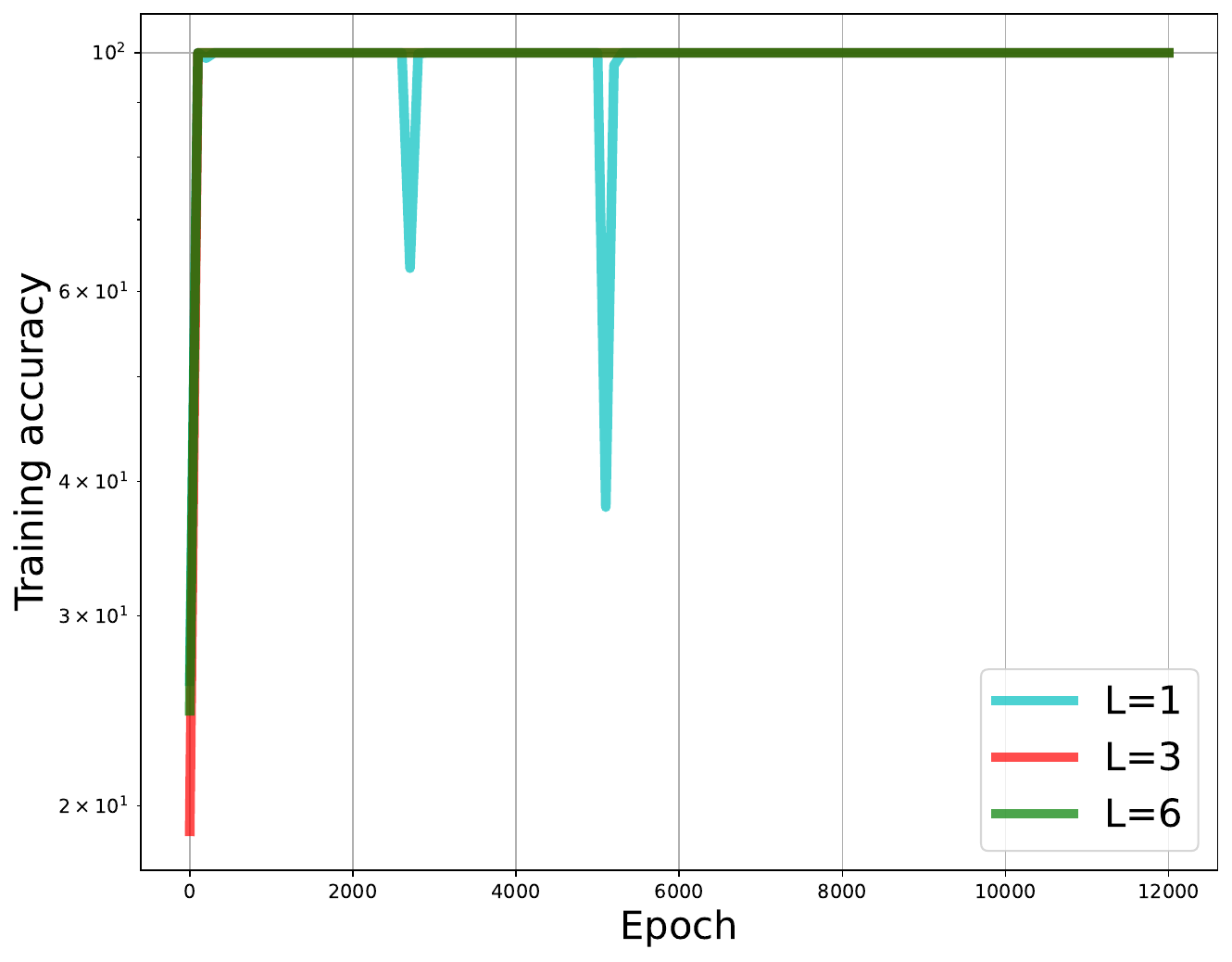}}
		\caption{Bias-free case: The performance of the NC metrics and training accuracy versus epoch with a 6-layer MLP backbone on an imbalanced subset of CIFAR10 for $L$-extended unconstrained feature model with $L=1,3,6$.} \label{fig:mlp1-cifar}
	\end{center}
\end{figure}
\begin{figure}[ht]
	\begin{center}
		\subfigure{
			\includegraphics[width=4.5cm, height=4cm]{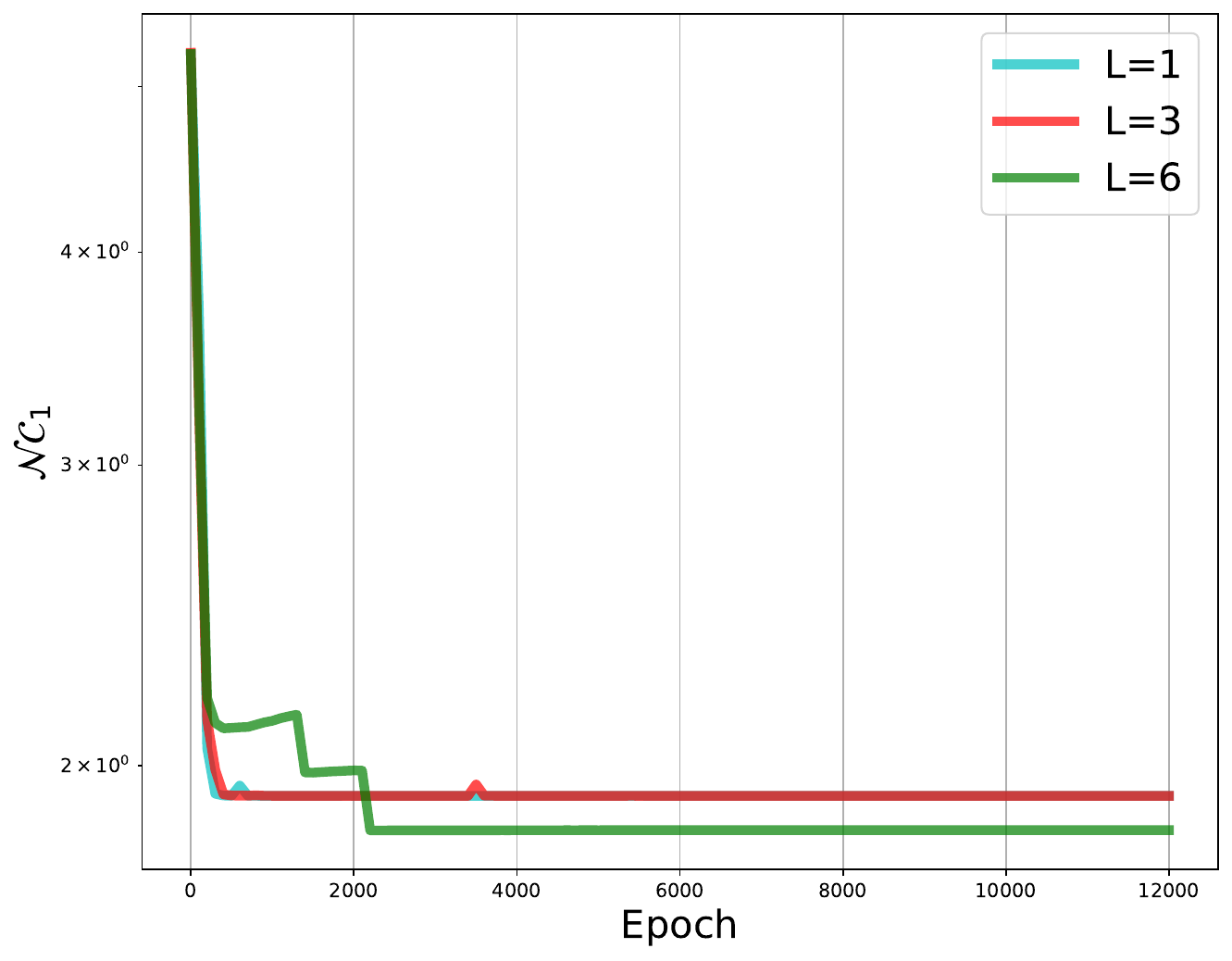}}
		\subfigure{
			\includegraphics[width=4.5cm, height=4cm]{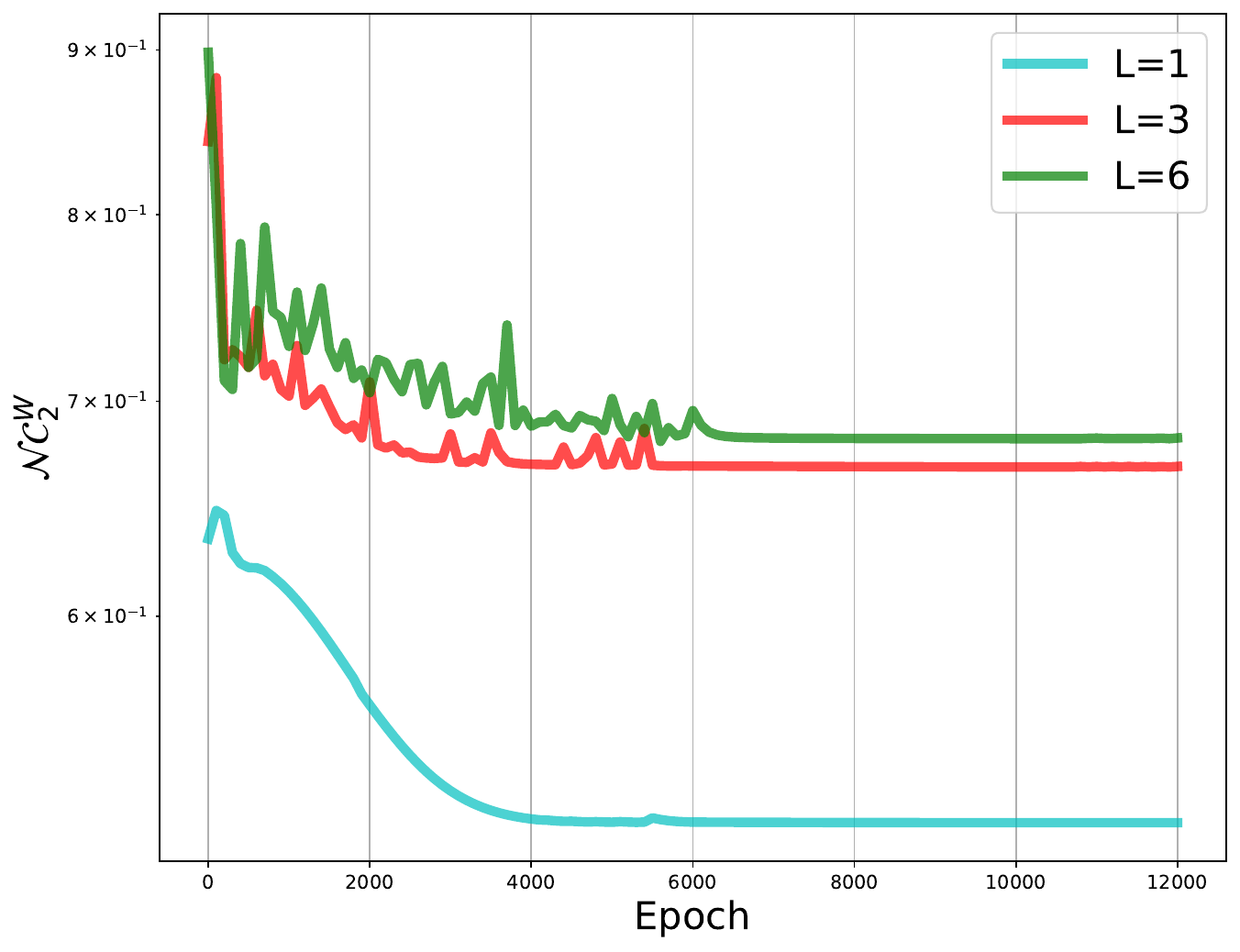}}
        \subfigure{
			\includegraphics[width=4.5cm, height=4cm]{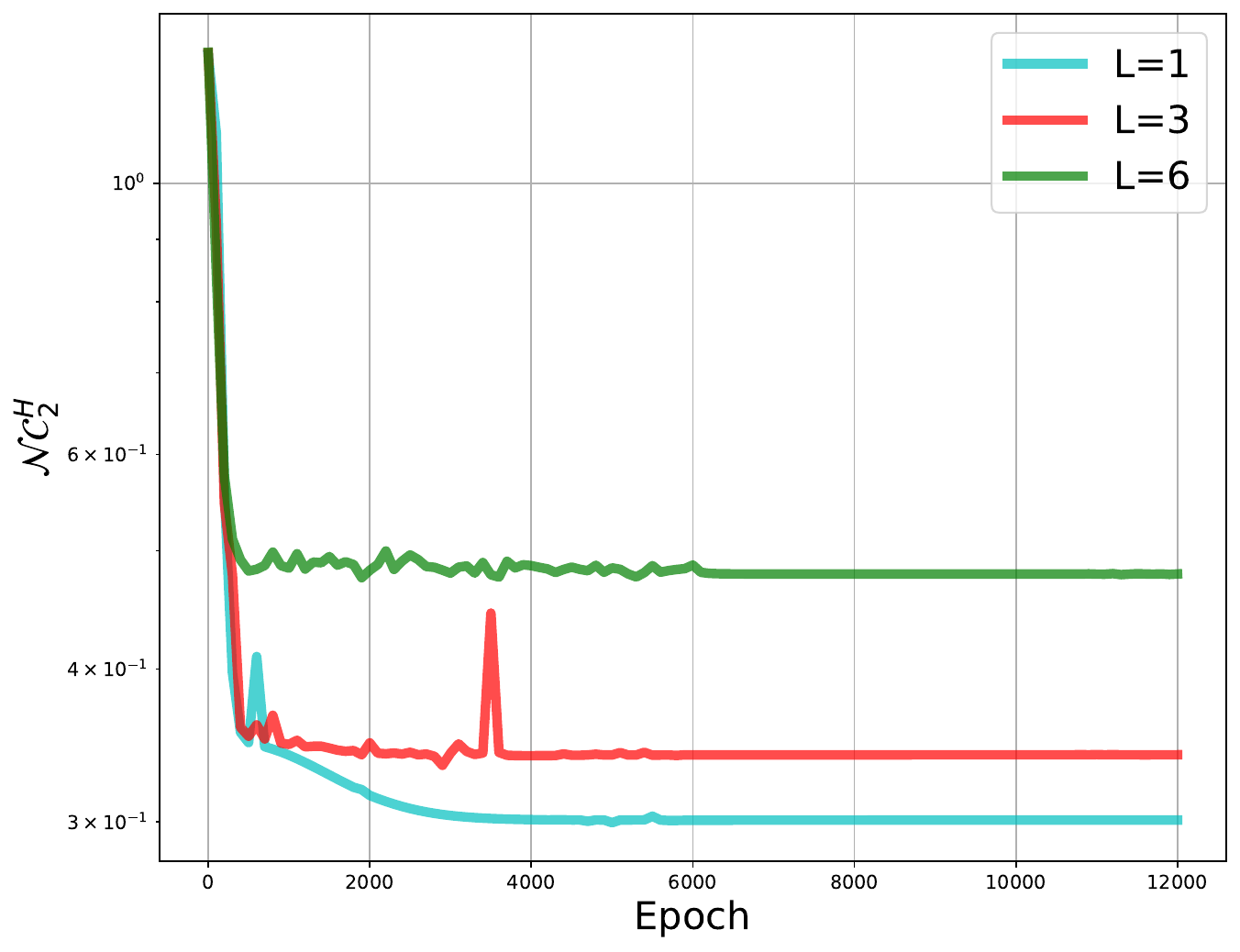}}
        \subfigure{
			\includegraphics[width=4.5cm, height=4cm]{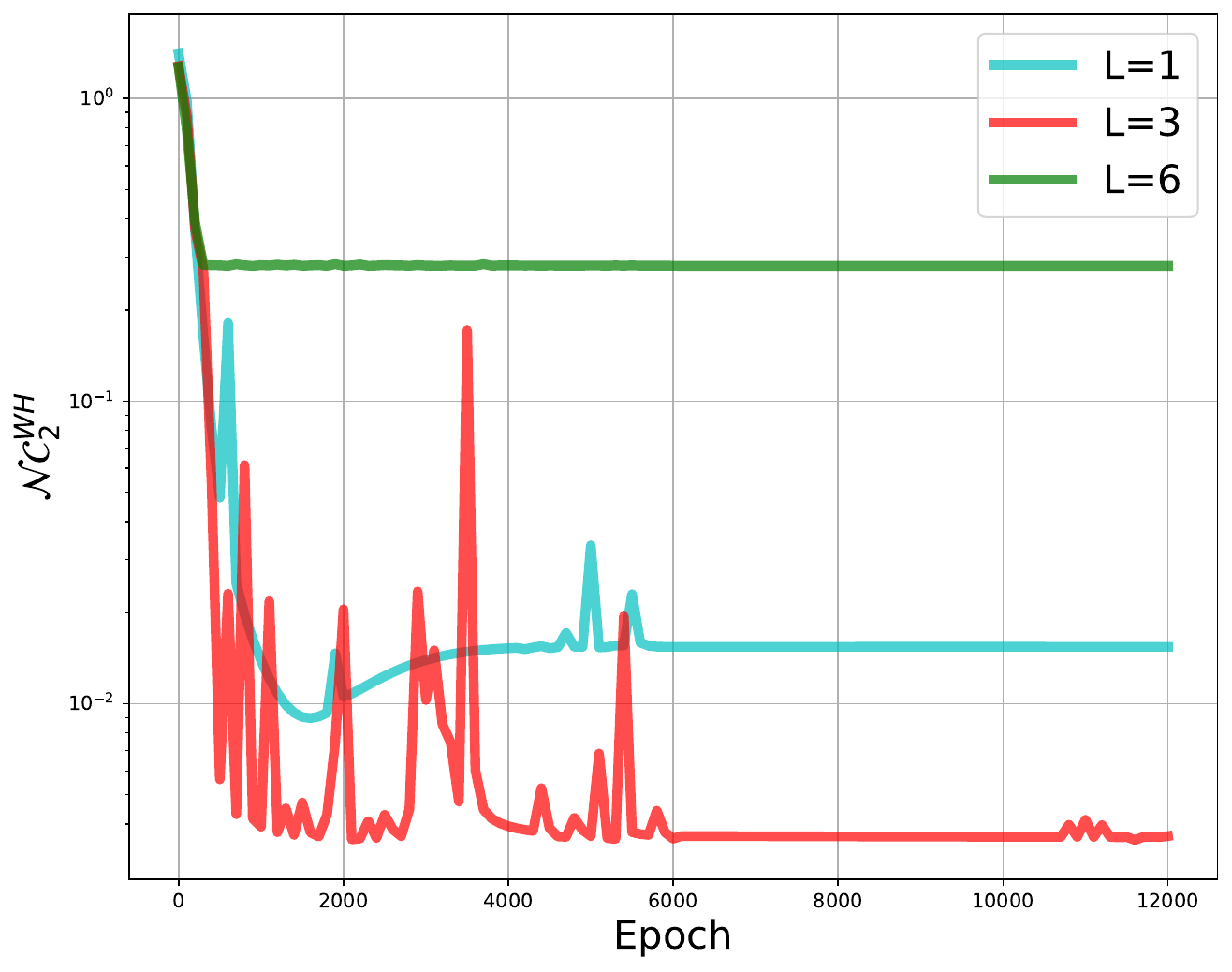}}
        \subfigure{
			\includegraphics[width=4.5cm, height=4cm]{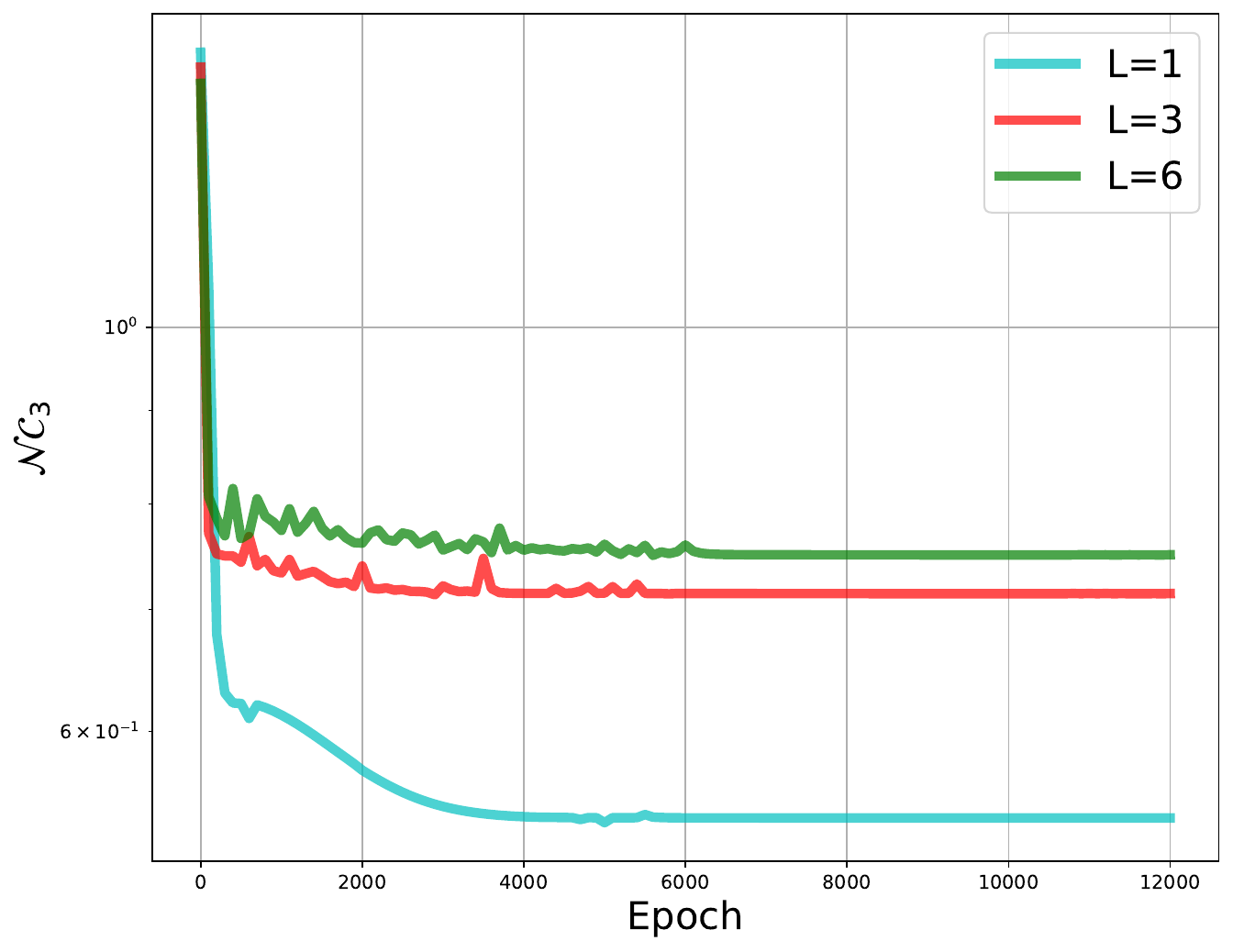}}
        \subfigure{
			\includegraphics[width=4.5cm, height=4cm]{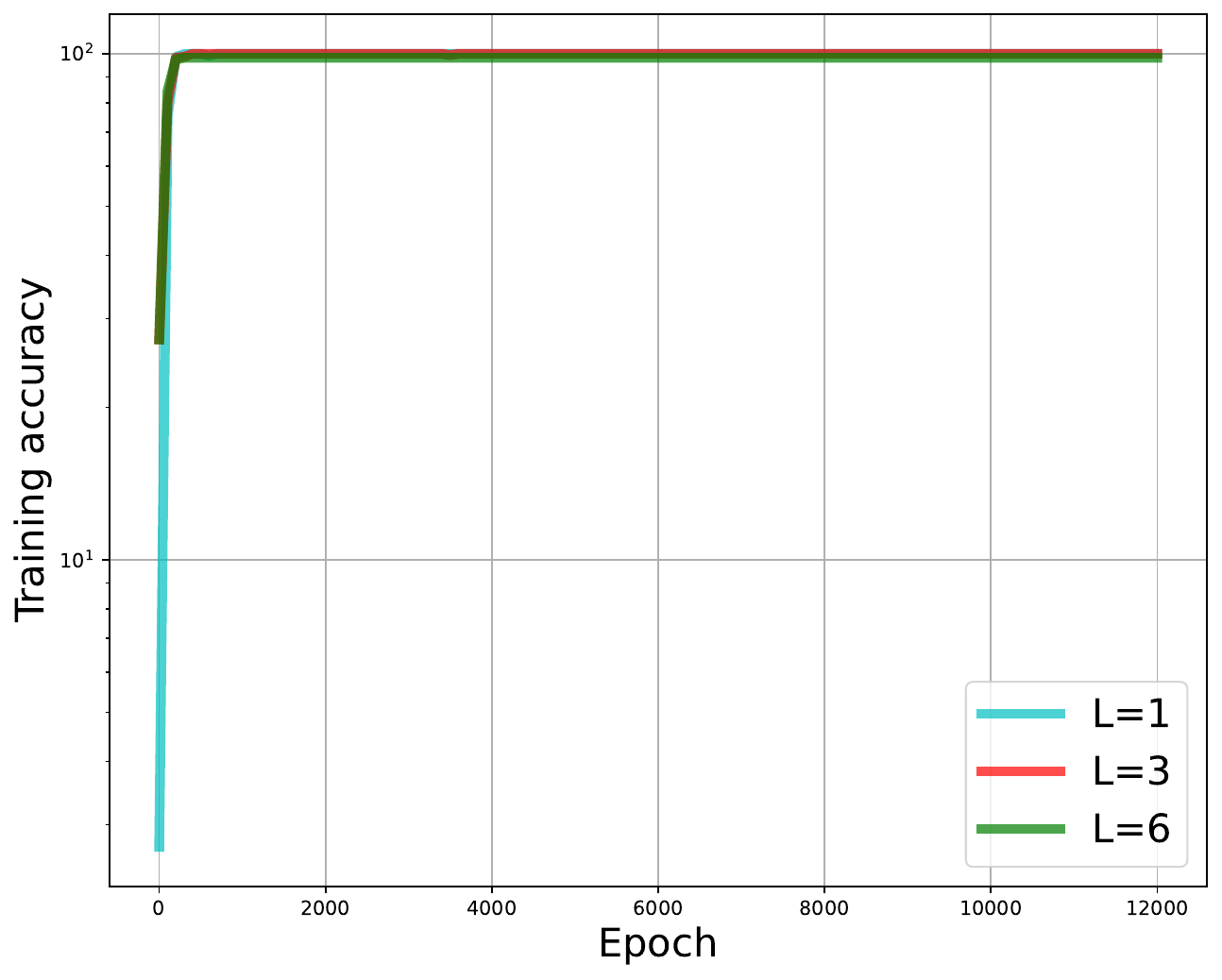}}
		\caption{Bias-free case: The performance of the NC metrics and training accuracy versus epoch with a 6-layer MLP backbone on an imbalanced subset of EMNIST for $L$-extended unconstrained feature model with $L=1,3,6$.} \label{fig:mlp1-emnist}
	\end{center}
\end{figure}
\section{Numerical Experiments on the Bias-Free Case}\label{appendix:numerical_bias-free}
The NC1 metric to evaluate the performance of neural collapse is the same as that in the bias case. The NC2 and NC3 metrics are defind as
    \begin{align*}
        NC^W_2=&\left\|\frac{(W_M\cdots W_1)(W_M\cdots W_1)^\top}{\|(W_M\cdots W_1)(W_M\cdots W_1)^\top\|_F}-\frac{\Upsilon^L_1}{\|\Upsilon^L_1
    \|_F}\right\|_F,\\
        NC^{H}_2=&\left\|\frac{\bar H^\top\bar H}{\|\bar H^\top\|_F}-\frac{\Upsilon_2}{\|\Upsilon_2
        \|_F}\right\|_F,\\
        NC^{WH}_2=&\left\|\frac{ W_M\cdots W_1\bar H}{\|W_M\cdots W_1\bar H\|_F}-\frac{\Upsilon_2}{\|\Upsilon_2
    \|_F}\right\|_F,
    \end{align*}
    where $\Upsilon_1,\Upsilon_2$ are defined in Theorem \ref{theo:bias_mult-layer}.\\
    Let $W= W_L\cdots W_1$ and $W^\top=\begin{bmatrix}
        w_1&\cdots&w_K
    \end{bmatrix}$, then
    \[NC_3=\left\|\begin{bmatrix}
        \frac{w_1}{\|w_1\|_2}&\cdots&\frac{w_K}{\|w_k\|_2}
    \end{bmatrix}-\begin{bmatrix}
        \frac{\bar h_1}{\|\bar h_1\|_2}&\cdots&\frac{\bar h_K}{\|\bar h_K\|_2}
    \end{bmatrix}\right\|_F.\]
Numerical experiments are illustrated in Figures \ref{fig:mlp1-cifar} and \ref{fig:mlp1-emnist} for CIFAR10 and EMNIST datasets, respectively.

\end{document}